\documentclass{article}

\usepackage{PRIMEarxiv}

\usepackage[utf8]{inputenc} % allow utf-8 input
\usepackage[T1]{fontenc}    % use 8-bit T1 fonts
\usepackage{hyperref}       % hyperlinks
\usepackage{url}            % simple URL typesetting
\usepackage{booktabs}       % professional-quality tables
\usepackage{amsfonts}       % blackboard math symbols
\usepackage{nicefrac}       % compact symbols for 1/2, etc.
\usepackage{microtype}      % microtypography
\usepackage{lipsum}
\usepackage{fancyhdr}       % header
\usepackage{graphicx}       % graphics
\graphicspath{{media/}}     % organize your images and other figures under media/ folder

\usepackage[utf8]{inputenc}
\usepackage[ruled,vlined]{algorithm2e}
\usepackage{geometry}
\usepackage{xcolor}
\usepackage{subcaption}
\usepackage{multirow}
\usepackage[flushleft]{threeparttable}
\usepackage{makecell}
\usepackage{hyperref}
\usepackage{amsmath}
\title{The COVMis-Stance dataset:\\ 
Stance Detection on Twitter \\
for COVID-19 Misinformation
}

\author{
  Yanfang Hou, Peter van der Putten, Suzan Verberne \\
  Leiden Institute of Advanced Computer Science \\
  Leiden University \\
  Leiden, the Netherlands\\
  \texttt{\{s.verberne, p.w.h.van.der.putten\}@liacs.leidenuniv.nl} \\
}

\begin{document}
\maketitle

\begin{abstract}
During the COVID-19 pandemic, large amounts of COVID-19 misinformation are spreading on social media. We are interested in the stance of Twitter users towards COVID-19 misinformation. However, due to the relative recent nature of the pandemic, only a few stance detection datasets fit our task. We have constructed a new stance dataset consisting of 2631 tweets annotated with the stance towards COVID-19 misinformation. In contexts with limited labeled data, we fine-tune our models by leveraging the MNLI dataset and two existing stance detection datasets (RumourEval and COVIDLies), and evaluate the model performance on our dataset. Our experimental results show that the model performs the best when fine-tuned sequentially on the MNLI dataset and the combination of the undersampled RumourEval and COVIDLies datasets. Our code and dataset are publicly available at \url{https://github.com/yanfangh/covid-rumor-stance}.
\end{abstract}

% keywords can be removed
\keywords{dataset \and stance detection \and misinformation \and COVID-19}

\section{Introduction}
In March 2020, the WHO declared the coronavirus (COVID-19) outbreak a global pandemic. We are also  experiencing a COVID-19 infodemic, in which  an excess of information is being created and shared to every corner of the world. In the early stages of the pandemic, misinformation surrounding COVID-19 was rampant due to a lack of knowledge about the virus itself and heated debates around how to best deal with it. It included false or misleading information and conspiracy theories towards the origin, scale, prevention, diagnosis, and treatment of the disease~\cite{COVID-19Misinformation}. Misinformation causes confusion, leads people to reject public health measures such as vaccination and masks, and promotes unproven treatments~\cite{office2021confronting}. 

With growing digitization, social media has become an important channel for information gathering and diffusion. Misinformation comes from a variety of sources, such as news sites, videos, and user posts. All of them can be easily shared on social media and further circulated and discussed by more users. Consequently, misinformation spreads rapidly on social media. At the time of posting, the veracity status of the information is usually not yet verified. Different social media users express different stances on its likely veracity and even share evidence supporting their views. Biber and Finegan~\cite{biber1988adverbial} define stance as the expression of a speaker’s standpoint and judgement towards a given proposition. These stances can be aggregated to measure public opinions and help determine the veracity of the rumors~\cite{zubiaga2018detection}. 

In this work\footnote{Research carried out by first author towards obtaining a MSc degree in computer science at LIACS, Leiden University}, we explore the stance expressed by tweets towards COVID-19 misinformation. Given a misinformation item and a tweet, our task is to classify each sentence pair into one of three categories: \textit{Favor}, \textit{Against} and \textit{Neither}. For instance, considering the following misinformation item and tweet, we can deduce from the tweet that the tweeter is likely in favor of the misinformation:

\begin{itemize}
    \setlength\itemsep{0em}
    \item[] \textbf{Misinformation}: 5G generates coronavirus in human skin cells.
    \item[] \textbf{Tweet}: The 5G technology breaks up the cells in human body to make the new Coronavirus.
\end{itemize}

Misinformation refers to wrong or misleading information, which is consistent with the definition given by the Oxford English Dictionary~\footnote{https://www.oed.com/}. Another term, rumor, is also used in this paper. According to the Oxford English Dictionary, a rumor is an unverified or unconfirmed statement or report in circulation. It may later be proven to be truth or misinformation, or keep unverified. In our work, we do not distinguish between misinformation and rumor, because they both have similar textual forms and they are computationally similar for stance detection. We also interchangeably use misinformation and rumor in this paper.

Previous stance detection datasets have been used to facilitate relevant research in various domains~\cite{mohammad2016semeval}~\cite{fnc1}~\cite{thorne-etal-2018-fact}, but stance detection datasets for targets relevant to COVID-19 are limited due to the recency of the pandemic. Glandt el al.~\cite{glandt2021stance} published a COVID-19 stance dataset, a collection of tweets annotated for stance towards controversial topics (e.g. \textit{Anthony S. Fauci, M.D.} and \textit{Stay at Home Orders}). The target in this dataset is usually an entity, while ours is described as a sentence. Hossain et al.~\cite{hossain2020covidlies} released the COVIDLies dataset to explore the stance of tweets towards COVID-19 misinformation. This task is the same as ours, but the dataset shows a great imbalance in class distribution, in which the majority of tweets have no stance. Considering the limited stance data available on COVID-19 rumors, we created our own dataset for stance detection, called COVMis-Stance~\footnote{https://github.com/yanfangh/covid-rumor-stance}. It comprises 2631 tweets annotated for stance towards COVID-19 misinformation.

Our first research question is: How do we use the external datasets for model training to detect stance in the COVMis-Stance dataset? Due to size of the data set, it is a challenge to obtain an effective model simply by fine-tuning the model on our dataset. Also, we want our model to have a certain degree of generalization ability. 

Therefore, we add a setting to our task that the training set and the test set have different sources. Specifically, we only use our dataset as the test set and do not use it to fine-tune the model. Instead, we fine-tune the model on datasets with similar tasks or domains. These datasets include the MNLI dataset~\cite{williams-etal-2018-broad} with a large number of instances, and two stance detection datasets with social media text, called RumourEval~\cite{gorrell-etal-2019-semeval} and COVIDLies~\cite{hossain2020covidlies}. we want to transfer knowledge from external datasets to the target domain.

To address this task, we build stance detection models using two architectures: Sentence-BERT (SBERT)~\cite{reimers-2019-sentence-bert} and Cross-Encoder~\cite{devlin-etal-2019-bert}. We use the models fine-tuned on the MNLI dataset and then fine-tune them on the RumourEval or COVIDLies datasets. In our task, we compare the performance of the models with different architectures and fine-tuned on different datasets. This leads to answering our second question is: Which of these models can achieve the highest quality in the stance detection task? 

Our contributions are four-fold:
\begin{itemize}
    \item We construct a COVID-19 stance dataset, called COVMis-Stance, consisting of 2631 tweets annotated with the stance towards 111 COVID-19 misinformation items, and make it publicly available. 
    \item We reproduce the stance detection experiments of Hossain et al.~\cite{hossain2020covidlies} for COVID-19 misinformation and obtain similar results.
    \item We establish stance detection models with the architecture of SBERT and Cross-Encoder. We found that Cross-Encoder outperforms SBERT in our task.
    \item We use the models fine-tuned on the MNLI dataset and then fine-tune them on the RumourEval or COVIDLies datasets. We evaluate the model performance on our dataset, and found that the model achieves the best result when fine-tuned sequentially on the MNLI dataset and the combination of the undersampled RumourEval and COVIDLies datasets. We also do a dataset ablation study to determine the extent to which each dataset contributes to solving the stance detection task.
\end{itemize}

The remainder of this paper is organized as follows. Section 2 presents the available datasets for stance detection and COVID-19 misinformation, and methods for stance detection and limited annotations. Section 3 describes how to construct the COVMis-Stance dataset and additional training datasets. Section 4 states the reproduction experiments of the COVIDLies paper~\cite{hossain2020covidlies}. Section 5 explains our stance detection methods. Section 6 describes the experimental setups and results of our task. Section 7 discusses the possible improvements, followed by our conclusions in Section 8. We add how to construct misinformation items, extract keywords, and implement data sampling strategy in Appendix~\ref{appendix}.

\section{Related Work}

In this section, we survey the existing work on data and methods relevant to our task. We present the available datasets for stance detection and COVID-19 misinformation. Also, we investigate stance detection methods and existing solutions to limited labeled data.

\subsection{Stance Detection Datasets}\label{std}
We summarize the existing stance detection datasets, shown in Table~\ref{std_data}. As these datasets are used for stance detection in various domains, they differ considerably in input texts, i.e. target and context. Among them, only COVIDLies has COVID-19 misinformation as its target.

\paragraph{SemEval-2016} The SemEval-2016 Task 6~\cite{mohammad2016semeval} presented a benchmark dataset for determining whether a tweet is in favor, against, or neither, of a given topic. Five topics are used in the dataset: \textit{Atheism}, \textit{Climate Change is a Real Concern}, \textit{Feminist Movement}, \textit{Hillary Clinton}, and \textit{Legalization of Abortion}. Compared to this task, our task is to determine the stance towards a rumor, which is usually an entire statement rather than a short entity or topic. 

\paragraph{FNC-1} The 2017 Fake News Challenge Stage 1 (FNC-1)~\cite{fnc1} is the task of determining the stance of the news body text towards the news headline. The labels are \textit{agree}, \textit{disagree}, \textit{discuss} (the body text discusses the same topic as the headline but does not take a position) and \textit{unrelated} (the body text discusses a different topic than the headline). In contrast, our task focuses on social media text rather than news. The tweet is usually shorter than the news body and contains many informal expressions. 

\paragraph{FEVER} Stance detection could be used for fact-checking. Thorne et al.~\cite{thorne-etal-2018-fact} presented the Fact Extraction and Verification (FEVER) shared task for classifying a claim as \textit{supported} or \textit{refuted} by Wikipedia evidence, or \textit{notenoughinfo} when the retrieved evidence is not relevant or informative. In addition to domain differences from our task, some claims in the FEVER dataset require the composition of evidence from multiple sentences, while each example in our dataset contains only one tweet for a given misinformation item.

\paragraph{RumourEval-2019}\label{rumoureval-2019} The RumourEval 2019 task~\cite{gorrell-etal-2019-semeval} shared a dataset about the stance expressed in the tweets in a conversation thread towards the rumor mentioned in the source tweet. The labels are \textit{support}, \textit{deny}, \textit{question} (the replying tweet requires additional evidence for the rumor veracity) and \textit{comment} (the replying tweet does not have a clear stance). The stance of each tweet towards the rumor is inferred from the contextual conversation. In contrast, the rumors in our dataset are not tweets. They are collected from news articles. Also, the tweets are independent and without conversational relationships.

\paragraph{Datasets related to COVID-19}~\label{covid_stance} Two types of targets in COVID-19 stance tasks attract great interest: controversial topics and rumors. Glandt et al.~\cite{glandt2021stance} released the COVID-19-Stance dataset, a collection of annotated tweets that express the stance towards four targets: \textit{Anthony S. Fauci, M.D.}, \textit{Keeping Schools Closed}, \textit{Stay at Home Orders}, and \textit{Wearing a Face Mask}. Hossain et al.~\cite{hossain2020covidlies} published a stance detection dataset, called COVIDLies, consisting of 6761 tweets with their annotated stance on COVID-19 misinformation.

\begin{table}\renewcommand\arraystretch{1.2}
  \centering
  \begin{threeparttable}
  \scalebox{0.9}{
  \begin{tabular}{lllll}
    \hline
    Dataset & Target & Context & Label Space \\
    \hline
    SemEval-2016~\cite{mohammad2016semeval} & Topic & Tweet & Favor, Against, Neither\\
    FNC-1~\cite{fnc1} & Headline & Article & Agree, Disagree, Discuss, Unrelated\\
    FEVER~\cite{thorne-etal-2018-fact} & Claim & Evidence & Supported, Refuted, NotEnoughInfo \\
    Rumoureval-2019~\cite{gorrell-etal-2019-semeval} & Tweet & Reply & Support, Deny, Query, Comment\\
    COVID-19-Stance~\cite{glandt2021stance} & Topic & Tweet & In-favor, Against, Neither\\
    COVIDLies~\cite{hossain2020covidlies} & Claim & Tweet & Agree, Disagree, No stance\\
    \hline
  \end{tabular}
  }
  \end{threeparttable}
  \caption{Datasets for stance detection}\label{std_data}
\end{table}

\subsection{COVID-19 Misinformation} 
A number of COVID-19 misinformation datasets have been released since the pandemic outbreak. Table~\ref{mis_data} lists a selection of available COVID-19 misinformation datasets. With the exception of COVIDLies, none of them are dedicated to stance detection. This means that if we use these datasets, we may need to collect tweets and annotate stance ourselves. Therefore, we focus specifically on three aspects of the misinformation datasets: the source from which the misinformation is collected, whether relevant tweets are included, and if not, whether the information is provided to facilitate the collection of relevant tweets. 

\begin{table}\renewcommand\arraystretch{1.2}
  \centering
  \begin{threeparttable}
  \scalebox{0.85}{
  \begin{tabular}{lcccccc}
    \hline
    \multirow{2}*{Dataset} & \multicolumn{4}{c}{Information Type} && \multirow{2}*{\makecell[c]{Relevant\\Tweets}} \\
    \cline{2-5}
    & News article & Fact-checking article & Claim & Post && \\
    \hline
    CoAID~\cite{cui2020coaid} & \checkmark  & \checkmark & \checkmark & - && \checkmark\\
    \hline
    FakeCovid~\cite{shahifakecovid}  & - & \checkmark & - & - && - \\
    \hline
    ReCOVery~\cite{zhou2020recovery} & \checkmark & - & - & - && \checkmark \\
    \hline
    CMU-MisCOV19~\cite{memon2020characterizing} & - & - & - & \checkmark && -\\
    \hline
    COVID-19FakeNews~\cite{patwa2021fighting} & - & - & - & \checkmark && -\\
    \hline
    COVIDLies~\cite{hossain2020covidlies} & - & - & \checkmark & - && \checkmark \\
    \hline
  \end{tabular}
  }
  \end{threeparttable}
  \caption{Datasets for COVID-19 misinformation. Includes the information types for each dataset and whether comprising the tweets relevant to misinformation.}\label{mis_data}
\end{table}

\paragraph{Fact-checking sites} Fact-checking websites are an important source for collecting fake news. Cui et al.~\cite{cui2020coaid} presented the CoAID dataset including true and fake news articles or claims related to COVID-19. Fake news is collected from the articles on fact-checking sites, and fake claims are from the WHO official website, the WHO Twitter account, and Medical News Today (MNT). Also, they provide relevant tweets by using the titles of news articles as search queries. However, such queries generate a limited number of related tweets due to the long titles, and they are biased to retrieve the tweets that support fake news. Therefore, if we use the CoAID dataset, it is necessary to extend the Twitter data. Shahi et al.~\cite{shahifakecovid} published a multilingual fact-checking news dataset called FakeCovid. In the process of collecting data, Snopes and Poynter are used as a bridge to get the links of fact-checking articles. Compared to CoAID, FakeCovid does not include the news articles as referred to in fact-checking articles. This means we cannot utilize any information from news articles to extend tweets.

\paragraph{News sites} Some studies do not resort to fact-checking sites but collect news directly from news sites. Zhou et al.~\cite{zhou2020recovery} proposed the ReCOVery dataset for COVID-19 news credibility research. They first determined unreliable news sites, relying on two sites that rate news media (NewsGuard and Media Bias/Fact Check), and then collected fake news from the sites using keywords. Also, tweets are obtained based on the URL of each news article. 

\paragraph{Social media} Some studies identify fake information from social media platforms. Memon et al.~\cite{memon2020characterizing} proposed CMU-MisCOV19, a diverse set of annotated COVID-19 tweets. They used pre-defined keywords and hashtags to retrieve tweets and annotated them according to their topics and veracity. Fake information is mentioned in tweets, similar to RumourEval-2019~\cite{gorrell-etal-2019-semeval}. This dataset is more appropriate for exploring the stance of a reply or quote tweet towards the rumor mentioned in a source tweet, while our task does not focus on conversational contexts. Patwa et al.~\cite{patwa2021fighting} collected COVID-19 news from both social media and fact-checking sites, and manually verified whether they are real or fake. The dataset consists of the content of each post with its annotated veracity, but no additional information.

\paragraph{Wikipedia} The COVIDLies~\cite{hossain2020covidlies} dataset described in Section~\ref{covid_stance} includes 86 misinformation items, which are collected from a Wikipedia article about COVID-19 misinformation~\cite{COVID-19Misinformation}. 

% \begin{table}\renewcommand\arraystretch{1.2}
%   \centering
%   \begin{threeparttable}
%   \scalebox{0.82}{
%   \begin{tabular}{lllll}
%     \hline
%     Dataset & Source & Info & Size & Lang \\
%     \hline
%     CoAID~\cite{cui2020coaid} & \makecell[l]{Fact-checking sites\\Official sites} & \makecell[l]{News article\\Fact-checking article}  & \makecell[l]{6201\\18.4\% fake}  & EN \\
%     \hline
%     FakeCovid~\cite{shahi2020fakecovid} & Fact-checking sites & \makecell[l]{Fact-checking article} & 7623  & \makecell[l]{Multi\\37.3\% EN}\\
%     \hline
%     ReCOVery~\cite{zhou2020recovery} & News sites & \makecell[l]{News article} & \makecell[l]{2029\\32.8\% fake} & EN\\
%     \hline
%     CMU-MisCOV19~\cite{memon2020characterizing} & Twitter & Tweet & \makecell[l]{4573\\31.1\% fake} & EN\\
%     \hline
%     COVID-19FakeNews~\cite{patwa2021fighting} & \makecell[l]{Twitter\\Fact-checking sites} & Post  & \makecell[l]{10700\\47.7\% fake} & EN\\
%     \hline
%     COVIDLies~\cite{hossain2020covidlies} & Wikipedia & Misinformation & 86 & EN \\
%     \hline
%   \end{tabular}
%   }
%   \end{threeparttable}
%   \caption{Datasets for COVID-19 misinformation}\label{mis_data}
% \end{table}

\subsection{Stance Detection Methods} 

This section describes the stance detection methods in three cases: (1) when the target is a topic; (2) the Fake News Challenge FNC-1 task; (3) fact-checking as stance detection.

\subsubsection{Topic} 

In the SemEval-2016 Task 6, Mohammad et al.~\cite{mohammad2016semeval} provided strong baseline results based on an SVM classifier with word n-grams and character n-grams features, but it performs poorly on unseen topics. How to generalize models across topics is an important research direction for stance detection. 

Reimers et al.~\cite{Reimers:2019:ACL} explored the use of contextualized word embeddings (ELMO and BERT) and topic information for argument classification. They use the concatenation of the topic and the sentence as the BERT input and then perform Softmax classification using the [CLS] token from the BERT output. The results show that it improves the macro F1 score by about 15\% points over the model without topic information. Based on this study, Reuver et al.~\cite{reuver-etal-2021-stance} investigated the extent to which the cross-topic model is topic-independent. They found the BERT model fluctuates on different topics. In addition, they analyzed the important lexical features through SVM classification, and conjectured that BERT models rely more on topic-specific features for stance detection than topic-independent lexical features.

\subsubsection{Fake News} 

A first step into the detection of fake news is understanding what a person or medium is saying about a particular topic or news items. So the first task of the Fake News Challenge (FNC) focussed on stance detection (FNC-1)~\cite{fnc1}: given a headline and a secondary text discussing the headline, the stance of the secondary text needs to be predicted (agrees/disagrees/discusses/unrelated).

Most participants addressed the problem by constructing neural networks and incorporating multiple hand-engineered features. The second-ranking system~\cite{atheneFNC}, called featMLP, is an ensemble of multi-layer perceptrons (MLP) with six hidden and a Softmax layer each. The input features consist of word unigrams, the similarity of word embeddings, topic models based on non-negative matrix factorization, latent Dirichlet allocation, and latent semantic indexing. Hanselowski et al.~\cite{hanselowski-etal-2018-retrospective} did a retrospective analysis for the methods and found that the most useful features are lexical features, followed by the topic model-based features. 

Since the FNC-1 competition, pretrained models have achieved great improvements in NLP tasks. Slovikovskaya et al.~\cite{slovikovskaya-attardi-2020-transfer} investigated the performance of transformer-based models on the FNC-1 task by exploiting two strategies: (1) Based on featMLP, they add the BERT sentence embeddings of two input sequences along with two similarity scores between them as model features. This results in a 2\% points increase in macro F1 and a 7\% points increase for the most difficult \textit{disagree} class; (2) The transformer-based models (BERT, RoBERTa, XlNet) 
are fined-tuned on the FNC-1 data, and they perform 11\%-16\% higher in macro F1 than the base featMLP. 

\subsubsection{Fact-checking as Stance Detection}
The FEVER task~\cite{thorne-etal-2018-fact} is commonly seen as a stance detection task. Specifically, the pipeline of the FEVER task usually consists of three components: document retrieval (select documents related to the claim), sentence selection (extract the most relevant sentences as evidence), and claim verification (determine whether the claim is supported or refuted by the evidence). The third component is about stance detection, so we mainly describe its solution.  

The winning system of the FEVER competition was proposed by Nie et al.~\cite{nie2019combining}. They framed the three-stage FEVER task as a similar semantic matching problem and proposed the Neural Semantic Matching Network (NSMN), a modification of ESIM~\cite{chen-etal-2017-enhanced}. For claim verification, the input sequences are a claim and a set of evidential sentences.

First, each input token consists of the following features: Glove and Elmo embeddings, WordNet embeddings, and semantic relatedness scores from the two upstream stages. Second, a bidirectional LSTM (BiLSTM) layer is used to encode each token. Then, a dot-product attention mechanism is utilized to obtain the aligned token representation. In addition to the original representation, the element-wise difference and product between the encoded and aligned representations are also combined to model complex interactions. Next, the model takes the upstream compound representation and keeps using BiLSTM to construct the representation. Finally, the max-pooling, along with their absolute difference and element-wise multiplication is used for classification. This system achieves a FEVER score of 64.0, about 2.3 times greater than the baseline result.

In recent years, many studies use pretrained models for the FEVER task. Soleimani et al.~\cite{soleimani2020bert} investigated the effect of BERT for sentence selection and claim verification, and finally achieves a FEVER score of 69.7. Jiang et al.~\cite{jiang2021exploring} take advantage of the T5 model for claim verification. T5~\cite{2020t5} is a sequence-to-sequence transformer-based model, pre-trained on a multi-task mixture of unsupervised and supervised datasets by reframing all NLP tasks into a unified text-to-text format. Given a claim and a set of candidate evidence, they fine-tune the model with the following input template: 
\begin{equation*}
    query: q\;sentence_1: s_1\;...\;sentence_L: s_L\;relevant:
\end{equation*}
where $q$ and $s_i$ are the claim and evidence sentences. $query$, $sentence_i$ and $relevant$ indicate that they are followed by the claim, evidence, and stance strings, respectively. The model will generate one of three tokens: \textit{supported}, \textit{refuted} and \textit{noinfo}. This system finally attains a FEVER score of 75.87.

\subsection{Methods for Limited Annotations}
The lack of large amounts of labeled data is the main challenge for COVID-19 related tasks. In previous work, two ideas are commonly used to address this problem, one is data augmentation and the other is transfer learning, i.e. transferring knowledge from a source setting to a different target setting. As both cover a wide range of topics, here we just describe some advanced methods or methods relevant to our task. 

\paragraph{Self-training} With a small manually labeled dataset, Miao et al.~\cite{miao-etal-2020-twitter} adopted self-training and knowledge distillation for data augmentation. Specifically, a teacher model is first trained with the manually labeled dataset, and then used to generate pseudo labels for the unlabeled data. After that, a student model is initialized with the identical architecture and parameters as the teacher model and trained with the union of manually and pseudo labeled data. This student model becomes a new teacher model and this process iterates over several times. The experiments show that the student model outperforms the teacher model by about 10\% points in terms of accuracy.

\paragraph{Intermediate tasks} One way to teach the model abilities is to fine-tune the model on a data-rich intermediate task before task-specific fine-tuning. Pruksachatkun et al.~\cite{pruksachatkun-etal-2020-intermediate} investigated the effect of various intermediate tasks by performing extensive experiments. They found that the intermediate tasks which require high-level inference and reasoning capabilities work best. The MNLI task is also used as an intermediate task in this study, which improves the accuracy by an average of 0.7\% points across 10 target tasks.

\paragraph{Prompt learning} Prompt or pattern-based training has emerged as an effective method of exploiting pretrained language models for few-shot learning. It reduces different NLP tasks into a masked language modeling problem, thus making better use of the knowledge encoded in the pretrained models. Hardalov et al.~\cite{hardalov-etal-2022-fewshot} explored the effect of this method on few-shot cross-lingual stance detection. The prompt in the stance task has the format:
\begin{itemize}
    \item [] \textit{[CLS]The stance of the following \underline{CONTEXT} is [MASK] the \underline{TARGET}.[SEP]}
\end{itemize}
The \textit{\underline{CONTEXT}} and the \textit{\underline{TARGET}} are replaced by the corresponding content of each example. For instance, given the context \textit{``I am so happy that Donald Trump lost the election."} and the target \textit{``Donald Trump"}, the input text is \textit{``[CLS]The stance of the following I am so happy that Donald Trump lost the election. is [MASK] the Donald Trump.[SEP]"}. The masked token is expected to be \textit{against}. In few-shot settings, this work transfers knowledge from English stance datasets to multi-lingual stance tasks, which substantially improves the model performance on multi-lingual datasets.

% Wright et al.~\cite{wright2020transformer} utilize the datasets from multiples sources for model training and combines their predictions on unseen dataset. They set different language models for each domain and the output probabilities of these models are mixed in four different methods: simple averaging, fine-tuning averaging, attention with a domain-classifier and a novel sample-wise attention mechanism based on transformer attention.

\paragraph{Multi-dataset learning} Mixing datasets from different domains or sources is often used to overcome resource limitations and improve the generalization ability of the model. Schiller et al.~\cite{schiller2021stance} presented a new stance detection benchmark that learns from multiple stance detection datasets. In the multi-dataset learning (MDL) setting, all datasets share the BERT architecture, but each has its own dataset-specific dense layer on top. Each dataset retains domain-specific information while sharing information through the encoder. With this framework, they fine-tune the model on all datasets simultaneously, showing a sizable improvement in the overall performance compared to the model trained on individual datasets. This study uses 10 stance detection datasets, while our task involves only four datasets. Thus, they have more diverse datasets and more complicated issues to handle. Another difference is that this study combines only datasets for the same task, while one of our training datasets is from a different task.

\section{Data}

In this section, we explain the collection and annotation of the COVMis-Stance dataset. Also, we describe three additional datasets used in our experiments: MNLI, RumourEval, and COVIDLies. We compare the datasets to ours and explain why we use them for training.

\subsection{COVMis-Stance}

\begin{figure}
\centering
\includegraphics[width=1.0\linewidth]{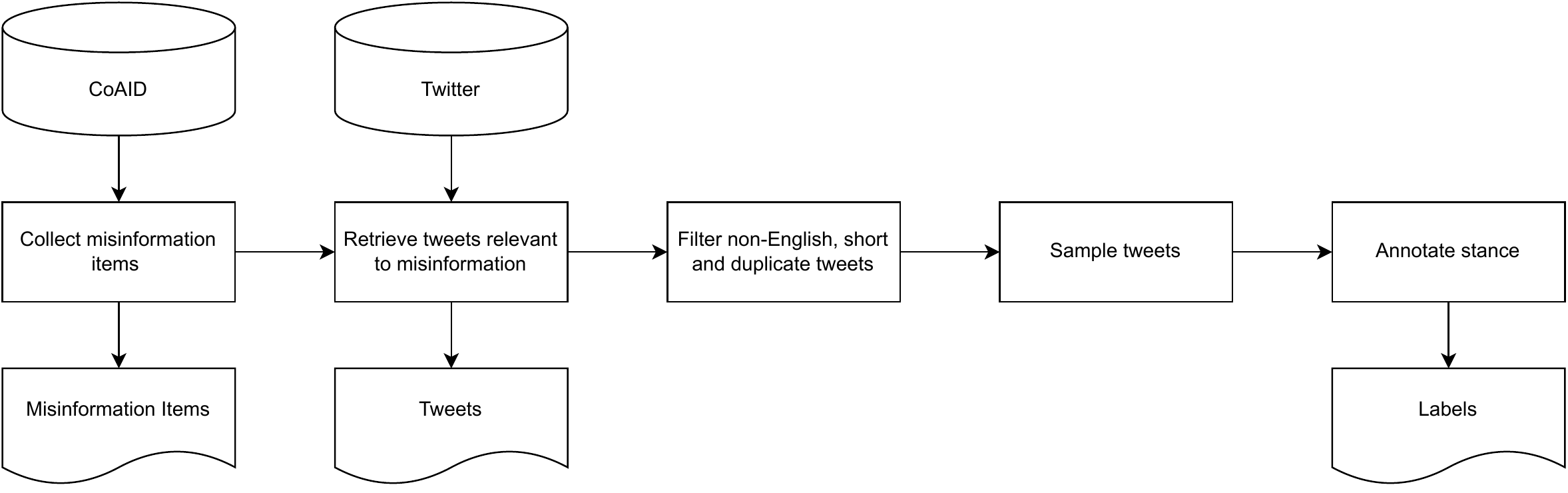}
\caption{Data construction pipeline of COVMis-Stance}\label{dc}
\end{figure}

Figure~\ref{dc} shows the data construction pipeline of COVMis-Stance. Our dataset consists of three components: misinformation, tweets, and labels. Misinformation comes from the CoAID dataset~\cite{cui2020coaid}, which includes COVID-19 fake news or claims on websites. To collect tweets related to our misinformation items, we build search queries using news titles, URLs of news or fact-checking articles, and news keywords, and fetch tweets matching these queries. After cleaning and sampling tweets, we annotate each tweet with the stance towards the misinformation item.

\paragraph{Misinformation}
We use the misinformation items in the CoAID dataset~\cite{cui2020coaid}. It consists of fake news and claims related to COVID-19. Fake news was collected from articles on fact-checking sites, and fake claims were from the WHO official website~\footnote{https://www.who.int/}, the WHO official Twitter account~\footnote{https://twitter.com/WHO} and Medical News Today (MNT)~\footnote{ https://www.medicalnewstoday.com/articles/coronavirus-myths-explored}. It provides abundant information for each misinformation item, including the titles, URLs, and contents of news articles, the titles and URLs of fact-checking articles, and the tweet IDs retrieved by news titles. For instance, an article~\footnote{https://healthfeedback.org/claimreview/human-dna-alone-does-not-produce-a-positive-result-on-the-rt-pcr-test-for-sars-cov-2/} on a fact-checking website refutes fake information in a news article~\footnote{https://pieceofmindful.com/2020/04/06/bombshell-who-coronavirus-pcr-test-primer-sequence-is-found-in-all-human-dna/}, titled \textit{``BOMBSHELL: WHO Coronavirus PCR Test Primer Sequence is Found in All Human DNA"}.

We sort the misinformation items based on the number of tweets retrieved by news titles and URLs, from highest count to lowest, and we select the top 200 items for our task. Each misinformation item is described as a sentence based on the titles of news articles or fact-checking articles. We preprocess the descriptions of some misinformation items to make them clear and specific, for example, splitting a compound misinformation item into multiple items. This is described in Appendix~\ref{mc} in detail.
% how to choose misinformation items

\paragraph{Tweet retrieval} To obtain tweets relevant to our misinformation items, we use the following queries to retrieve tweets: 
\begin{itemize}
    \item Titles of news articles: CoAID has provided the tweet IDs retrieved by news titles, so we directly fetch these tweets by IDs.
    \item URLs of news articles and fact-checking articles: The URLs are from CoAID. If the fake news is from Twitter, we will get the tweet itself and retrieve the tweets containing the URL of this tweet.
    \item Keywords: The keywords are manually extracted from each misinformation item. Appendix~\ref{ke} describes how to construct the keywords specific to the rumors.
\end{itemize}
Since the Twitter Search API only searches recent tweets published in the past seven days, we use a tool called Snscrape to retrieve tweets. Snscrape~\footnote{https://github.com/JustAnotherArchivist/snscrape} is a public, free scraper for social networking services (SNS). It supports the search service on Twitter and returns the discovered items. After obtaining the tweet IDs, we fetch the tweets with detailed information through Twarc~\footnote{https://github.com/DocNow/twarc}, which is a command-line tool and Python library for collecting and archiving Twitter JSON data via the Twitter API.

\paragraph{Data cleaning} We found that a large number of tweets with different IDs have the same contents. Also, some tweets are very short. Before sampling for annotation, we clean the data by following the steps below.
\begin{enumerate}
    \item Some tweets are associated with multiple misinformation items, so the dataset contains duplicate tweet IDs. We remove duplicate tweets and randomly keep one in the dataset.
    \item Remove non-English tweets based on the language identification of each tweet.
    \item Exclude the tweets with fewer than 10 words.
    \item Exclude the tweets with the same contents. we fit a TF-IDF vectorizer by using the whole tweet data and then compute the cosine similarity for each pair of tweets. If the similarity score is greater than the threshold (80\%), we assume both tweets are the same. We cluster the same tweets into one group and randomly keep one tweet for each group. 
    \item Exclude the misinformation items with fewer than 24 relevant tweets.
\end{enumerate}

\paragraph{Data sampling}~\label{sampling} Our sampling strategy is based on two principles: (1) keeping a good balance in the number of examples between different misinformation items; (2) covering examples of different query types. Here, we consider fact-checking URLs and news URLs as different query types to ensure the number of \textit{Against} examples. Specifically, we sample 24 tweets for each misinformation item. In these 24 tweets, we randomly select 6 tweets from each query type. However, some query types might not have enough tweets. In this case, if we still need $m$ examples to reach 24, we randomly sample $m$ tweets from the rest of the tweets. According to this strategy, we calculate the probability of being selected for each example, and randomly select examples from the entire set based on these probabilities. Therefore, the number of examples for each misinformation item is not exactly the same, but almost the same. Appendix~\ref{ids} describes in detail how to calculate the probability of being selected for each example.

\paragraph{Annotation}
Our annotation task is to annotate each tweet-rumor pair into one of three labels: \textit{Favor}, \textit{Against}, and \textit{Neither}. The core instructions given to annotators for determining stance are shown below:

\begin{itemize}
    \item Favor: The tweet is in favor of the misinformation or promotes the dissemination of fake news articles.
    \item Against: The tweet denies the misinformation or promotes the propagation of fact-checking articles.
    \item Neither: Neither of the above. It is usually one of these cases: (1) The tweet is unrelated to the misinformation; (2) The tweet questions the veracity of the misinformation; (3) The tweet has no clear stance for the misinformation.
\end{itemize}

In our annotation task, one type of examples is automatically labeled. If the tweet is retrieved by the URL of a fact-checking article, we directly annotate it as \textit{Against}. The rest of the examples were manually annotated by two annotators. Both annotators are master students studying AI-related programs, and English is their teaching language. For every 12 misinformation items, we calculate the inter-rater agreement on the annotated examples, discuss the examples on which the two annotators disagree, summarize the existing problems, and improve the annotation guideline. If we still cannot obtain a consistent result, we will ask a third person for advice.

\begin{table}\renewcommand\arraystretch{1.25}
  \centering
  \begin{threeparttable}
  \scalebox{0.8}{
  \begin{tabular}{p{2.0cm}p{1.5cm}p{13cm}}
    \hline
    Query & Label & Example \\
    \hline
    Title & Favor & \textbf{Misinformation:} Shanghai government officially recommends Vitamin C for COVID-19.\newline \textbf{Tweet:} Shanghai Government Officially Recommends Vitamin C for COVID-19\newline US Gov./Media Still Ignoring Solution https://t.co/n40linrF8z \\
    \hline
    URL & Against & \textbf{Misinformation:} US Education Secretary Betsy DeVos said children are stoppers of coronavirus. \newline \textbf{Tweet:} Playing politics with the lives of children. We tell them to listen to and trust adults. What fools we are. https://t.co/DXXM3CFN3D \\
    \hline
    Keywords & Neither & \textbf{Misinformation:} Queen Elizabeth II tests positive for coronavirus. \newline \textbf{Tweet:} Prince Charles tests positive for Coronavirus; Queen Elizabeth II too under quarantine\newline Prince of Wales displayed mild symptoms of Covid-19 and is otherwise in good health.\newline from Jagran Josh \\
    \hline
  \end{tabular}}
  \end{threeparttable}
  \caption{Examples of COVMis-Stance}\label{our_examples}
\end{table}

\paragraph{Data statistics}
We annotated a total of 2631 tweets towards 111 misinformation items, of which 604 \textit{Against} tweets are automatically annotated. The Cohen’s Kappa score for the manual annotation is 0.67, indicating substantial agreement between annotators (0.61-0.80). Some examples are presented in Table~\ref{our_examples}. The distribution of labels and query types are shown in Table~\ref{our_stats}. The \textit{Neither} class accounts for the smallest proportion, and the $Favor$ class has more tweets than $Against$. On the other hand, more tweets are retrieved by keywords than URLs. Only a small number of tweets are obtained by titles. We observe that the examples sourced by different query retrievals show a different class distribution. 83\% of title examples are \textit{Favor}, 59\% of URL examples are \textit{Against}, and 57\% of keywords examples are \textit{Favor}. 

\begin{table}\renewcommand\arraystretch{1.2}
  \centering
  \begin{threeparttable}
  \scalebox{0.9}{
  \begin{tabular}{cccc|c}
    \hline
    & Favor & Against & Neither & Total \\
    \hline
    Title & 195 & 38 & 3 & 236 (9.0\%)\\
    URL & 336 & 646 & 104 & 1086 (41.3\%)\\
    Keywords & 745 & 363 & 201 & 1309 (49.7\%)\\
    \hline
    Total & 1276 (48.5\%) & 1047 (39.8\%) & 308 (11.7\%) & 2631\\
    \hline
  \end{tabular}
  }
  \end{threeparttable}
  \caption{Statistics of COVMis-Stance} \label{our_stats}
\end{table}

\subsection{Additional Training Datasets}

\paragraph{MNLI} The Multi-Genre Natural Language Inference (MNLI) dataset~\cite{williams-etal-2018-broad} is a large collection of hypothesis/premise pairs annotated with their relationships (entailment, contradiction, and neutral). If the premise $p$ entails the hypothesis $h$, then a human reading $p$ would infer that $h$ is most likely true. For example, given the premise ``A turtle danced", we could infer that ``A turtle moved", but the reverse is not certain. Table~\ref{mnli_stats} shows the class distribution of the MNLI dataset. We see that the training set has a nearly equal number of examples on three labels. Similar to the NLI task, our stance detection task also focuses on the relationship between two sentences. We also have a similar label space to the NLI task. In addition, MNLI is a cross-domain dataset, covering ten distinct genres of written and spoken English. Thus, the MNLI dataset can provide a large and diverse set of training instances.

\begin{table}\renewcommand\arraystretch{1.2}
  \centering
  \begin{threeparttable}
  \scalebox{0.9}{
  \begin{tabular}{ccccccc}
    \hline
    & Entailment & Contradiction & Neutral & - & Total \\
    \hline
    Train & 130899 & 130903 & 130900 & 0 & 392702\\
    Dev-matched & 3479 & 3213 & 3123 & 185 & 10000\\
    Dev-mismatched & 3463 & 3240 & 3129 & 168 & 10000\\
    \hline
  \end{tabular}}
  \end{threeparttable}
  \caption{Statistics of MNLI. The distribution of labels in the training (\textit{train}) and development (\textit{dev}) set. \textit{Dev-matched} represents the development set has the same source with the training set, whereas \textit{Dev-mismatched} indicates not. The fifth column \textit{-} represents no label.}\label{mnli_stats}
\end{table}

\paragraph{RumourEval} The RumourEval-2019 dataset~\cite{gorrell-etal-2019-semeval} is about the stance expressed in tweets in a conversation thread towards the rumor mentioned in the source tweet. The labels are \textit{support}, \textit{deny}, \textit{query} and \textit{comment}. Both RumourEval and COVMis-Stance are about the stance in tweets towards the rumors, but differ in four aspects: 
\begin{itemize}
    \item Sources of rumors: Our rumors are from news articles, while the rumors in RumourEval are from tweets. As a result, the source tweets also take a stance towards the rumors, so we exclude the examples where the source tweets deny the rumors.
    \item Relations of tweets: The tweets in a conversation have a reply relationship for RumourEval, while our tweets are independent.
    \item Label space: RumourEval defines four labels. It has one more than ours, but these labels do not contradict ours. Thus we directly correspond \textit{Query} and \textit{Comment} to \textit{Neither} in our task. 
    \item Class distribution: The RumourEval dataset has a different class distribution from ours. This can be seen from Table~\ref{std_stats}. There is a class imbalance in the RumourEval dataset. Majority examples belong to the \textit{Neither} class, and the \textit{Favor} examples are more than \textit{Against}.
\end{itemize}

\paragraph{COVIDLies}\label{covidlies} The COVIDLies dataset~\cite{hossain2020covidlies} is a collection of annotated tweets that contain the stance towards a set of misinformation items. Each tweet-rumor pair is classified into one of three labels: \textit{Agree}, \textit{Disagree} and \textit{No Stance}. The statistics of COVIDLies v0.2 are shown in Table~\ref{std_stats}. The label distribution is heavily skewed to \textit{No stance}, and the $Agree$ tweets have a higher proportion than $Disagree$. Both COVIDLies and our dataset have rumors as targets and social media text, but differ greatly in the class distribution. This is caused by different strategies of tweet selection. For COVIDLies, Hossain et al. measure the similarity between tweets and misinformation items using BERTScore described in Section~\ref{bertscore}, and select the 100 most similar tweets for each misinformation item. They suggest that the lack of fine-tuning on tweets might degrade the BERTScore performance, leading to a high percentage of \textit{No Stance} tweets. In addition, there might be a bias in BERTScore to rate \textit{Agree} examples higher. By contrast, we associate tweets with misinformation through URLs and keywords, which may improve their matching.

% The authors suggest that this might be the result of tweet selection strategy. The tweets are from the collection of the COVID-19 related tweets. They use the BERTScore to measure the similarity on misconception-tweet pairs and select the 100 most similar tweets for annotation. However, Lack of fine-tuning on tweets might degrade the performance of BERTScore, and there might be a bias in BERTScore to score the \textit{Agree} examples higher.

% covidlies: 400/class num; 
% rumoureval: 550/class num

\begin{table}\renewcommand\arraystretch{1.2}
  \centering
  \begin{threeparttable}
  \scalebox{0.9}{
  \begin{tabular}{lcccc}
    \hline
    & Total & Favor & Against & Neither\\
    \hline
    RumourEval & 7730 & 796 (10.3\%) & 519 (6.7\%) & 6415 (83.0\%) \\
    COVIDLies v0.2 & 8937 & 738 (8.3\%) & 366 (4.1\%) & 7833 (87.6\%) \\
    \hline
  \end{tabular}}
  \end{threeparttable}
  \caption{Statistics of RumourEval and COVIDLies. Includes the number and percentage of examples in each label.} \label{std_stats}
\end{table}

\subsection{Data Examples}
Table~\ref{train} presents the \textit{Favor} examples of the above datasets. In the MNLI dataset, the premise seems to be more specific, while the hypothesis is more general. In the COVIDLies or our dataset, a misinformation item is usually a sentence that roughly summarizes the content of a rumor, while the tweet is usually longer and probably describes the event in more detail. Such difference is not obvious in the RumourEval dataset, because in the conversational context, the replying tweet is a response to the previous tweet rather than a restatement. 

\begin{table}\renewcommand\arraystretch{1.2}
  \centering
  \begin{threeparttable}
  \scalebox{0.8}{
  \begin{tabular}{{p{2.5cm}p{2.5cm}p{12cm}}}
    \hline
    Dataset & Label & Example \\
    \hline
    MNLI & Entailment & \textbf{Hypothesis:} Industry experts and federal agencies were involved in the study. \newline \textbf{Premise:} During the course of the study, a literature search was conducted, industry experts and practitioners were consulted, and federal agencies were Valueadded of Design surveyed.  \\
    \hline
    RumourEval & Support & \textbf{Source:} \#CharlieHebdo killers shot dead by police \newline \textbf{Reply:} @username you the fastest for breaking news...  \\
    \hline
    COVIDLies & Agree & \textbf{Misinformation:} Coronavirus is genetically engineered. \newline \textbf{Tweet:} Coronavirus was invented by the US to even out its economic game with China. Heard at a family gathering. https://t.co/hgKj9naWJX \\
    \hline
    COVMis-Stance & Favor &\textbf{Misinformation:} Charles Lieber, a Harvard University professor who was arrested, had a connection to the new coronavirus.\newline \textbf{Tweet:} Harvard University professor in trouble.
    Charles Lieber, charged with lying over his ties to China, received US\$50,000 a month and US\$1.5 million to start a research lab in Wuhan, prosecutors allege
    \#coronavirus \#WuhanPneumonia \#COVD19 
    https://t.co/25FNLXJ4Aj  \\
    \hline
  \end{tabular}}
  \end{threeparttable}
  \caption{Examples of the \textit{Favor} class}  \label{train}
\end{table}

\section{Reproduction Experiments}
In this section, we reproduce the stance detection experiments from the COVIDLies paper~\cite{hossain2020covidlies}, because this work also aims to detect stance in tweets towards COVID-19 misinformation. For each misinformation item and tweet pair, the task of this work is to predict whether the tweet \textit{Agree}, \textit{Disagree}, or takes \textit{No stance} with respect to the misinformation.

\subsection{Methods}
Hossain et al. frame the stance detection task as a NLI problem, mapping the tweet to the premise, the misinformation to the hypothesis, and the \textit{Agree}, \textit{Disagree} and \textit{No Stance} to \textit{Entailment}, \textit{Contradiction} and \textit{Neither} respectively. They train the models on the MNLI dataset and then evaluate their performance on the COVIDLies dataset. The methods are described as follows.

\paragraph{Logistic regression} 
Two features are separately used for logistic classification: (1) concatenating the unigram and bigram TF-IDF vectors of both sentences; (2) concatenating the average Glove embeddings of each sentence.

\paragraph{Sentence-BERT}~\label{sbert}
Reimers et al.~\cite{reimers-2019-sentence-bert} proposed Sentence-BERT, which adds a pooling layer to the output of BERT to obtain the representation of each sentence. The average pooling is used in this work. Given a tweet and a misinformation item, they use a Siamese BERT network to obtain the sentence representation $u$ and $v$ respectively. Then they concatenate $u$ and $v$ with the element-wise difference $|u-v|$ as features for Softmax classification. The classifier output $o$ is the probability that the input example belong to each of three labels, given by
\begin{equation}
    o=Softmax(W_t(u,v,|u-v|))
\end{equation}
$W_t$ is the trainable weights with a size of $3n\times k$, where $n$ is the dimension of sentence embeddings and $k$ is the number of labels.

\paragraph{BERTScore and Sentence-BERT}~\label{bertscore}
This is a two-step method. It first classifies whether each tweet-rumor pair is relevant. \textit{Relevant} refers to the tweet either agrees or disagrees with the rumor. Then it determines whether the pair \textit{Agree} or \textit{Disagree}. In the first step, it uses BERTScore to measure the relevance of each pair and classifies the pairs with high BERTScores as \textit{Relevant}. Then the Sentence-BERT model is used to determine \textit{Agree} or \textit{Disagree} in the second step.

BERTScore~\cite{zhang2019bertscore} is originally proposed to evaluate the quality of generated text relative to the gold references. Given a candidate sentence $\hat{x}$ and a reference sentence $x$, it consists of two components to obtain the similarity score of two sentences: (1) it computes a cosine similarity score for each token in the candidate sentence with each token in the reference sentence using BERT embeddings; (2) The BERTScore includes recall (R), precision (P) and F1. It matches each token in $x$ to the most similar token in $\hat{x}$ to calculate recall, and each token in $\hat{x}$ to the most similar token in $x$ to calculate the precision. The scores are given by: 
    \begin{align}
        R&=\frac{1}{|x|}\sum_{x_i\in x}\max_{\hat{x}_j\in\hat{x}} x_i^T\hat{x}_j\\
        P&=\frac{1}{|\hat{x}|}\sum_{\hat{x}_j\in\hat{x}}\max_{x_i\in x}x_i^T\hat{x}_j\\
        F_1&=2\cdot\frac{P\cdot R}{P+R}
    \end{align}
where the recall is the average matching score over all tokens in $x$, and the precision is the average matching score over all tokens in $\hat{x}$. The F1 score is used in the experiments.

\subsection{Experiments}

\paragraph{Models} 
We reproduced the results of five stance detection models. All models are trained on the MNLI dataset, and the COVIDLies dataset is used as the test set to evaluate the performance of these models. Among them, we retrain the linear models on the MNLI dataset (1 and 2 of the following models). Due to limited GPU resources, we do not retrain the SBERT models, but use the trained models from the original study for prediction (3, 4, and 5 in the following models).

\begin{enumerate}
    \item Linear, Bag-of-Words: The logistic classifier with the concatenation of unigram and bigram TF-IDF vectors for each sentence.
    \item Linear, Avg. GloVe: The logistic classifier with the concatenation of the average Glove embeddings for each sentence.
    \item SBERT: The model is initialized with the BERT-Base-Cased~\cite{devlin-etal-2019-bert} weights and fine-tuned on the MNLI dataset with the architecture of Sentence-BERT.
    \item SBERT (DA): Compared to SBERT, the model is initialized with the COVID-Twitter-BERT~\cite{muller2020covid} weights. COVID-Twitter-BERT is a BERT-large model, further pretrained on a large corpus of COVID-19 related tweets. The suffix \textit{DA} is appended to models that use COVID-Twitter-BERT, indicating domain-adaptive.
    \item BERTScore (DA) + SBERT (DA): The COVID-Twitter-BERT model is used to compute the BERTScore of each tweet-rumor pair. The BERTScore threshold is 0.4 for distinguishing between relevant and irrelevant examples. The SBERT(DA) model is then applied to the relevant examples to classify between \textit{Agree} and \textit{Disagree}.
\end{enumerate}

\paragraph{Data} 
The test set in the original experiments is COVIDLies v0.1, while we use the second version of COVIDLies, i.e. COVIDLies v0.2. Table~\ref{covidlies_stats} shows the label distributions of both datasets. COVIDLies v0.2 has the same source as COVIDLies v0.1, but with more instances. They also have similar label distributions. Another difference is that in the COVIDLies v0.2 dataset we use, username mentions in each tweet are replaced with @username due to Twitter's privacy policy. 

\begin{table}\renewcommand\arraystretch{1.2}
  \centering
  \begin{threeparttable}
  \scalebox{0.9}{
  \begin{tabular}{lcccc}
    \hline
    & Total & Agree & Disagree & No Stance\\
    \hline
    COVIDLies v0.1 & 6761 & 670 (9.9\%) & 343 (5.1\%) & 5748 (85.0\%) \\
    COVIDLies v0.2 & 8937 & 738 (8.3\%) & 366 (4.1\%) & 7833 (87.6\%) \\
    \hline
  \end{tabular}}
  \end{threeparttable}
  \caption{Statistics of COVIDLies v0.1 and v0.2. Includes the number and percentage of examples in each label.} \label{covidlies_stats}
\end{table}

\paragraph{Metrics}~\label{metric} Three metrics are used for evaluation per label: precision, recall and F1 score. The macro average is used to evaluate the overall performance of the models, which is the unweighted mean of the metrics of each label.
\begin{itemize}
    \item Precision: The fraction of true positives among the instances predicted to be positive, given by $P=tp/(tp+fp)$,  where $tp$ is the number of true positives and $fp$ the number of false positives.
    \item Recall: The fraction of true positives among the instances with positive labels, given by $R=tp/(tp+fn)$, where $fn$ is the number of false negatives.
    \item F1: This is calculated as the harmonic mean of precision and recall, given by $F_1=2\cdot(P\cdot R)/(P+R)$.
\end{itemize}

% he original paper displays the results on CovidLies v0.1, whereas we use the second version of CovidLies dataset.

\subsection{Results}
Table~\ref{lies_repr} presents the original and reproduction results of stance detection on the COVIDLies dataset. Despite the difference in test data size, our reproduction results are similar to those of the original study. Compared to the original results, 36 of the 60 metric values decrease and 24 metrics increase. For these 60 metrics, 49 of them differ from the original results with no more than 3.0\% points. The linear model with Glove embeddings has the largest difference in \textit{Agree} F1 from the original result, 5.2\% points lower (F1 = 20.5 vs 25.7). The score differences are only the result of the test set differences and are not related to models, since the models we use have the same parameters as those in the original study. This is because the linear models are deterministic and do not involve any randomness, and the SBERT models are from the original study. On the other hand, we observe that four of the five models decrease by 0.1\%-3.3\% points in macro F1 after expanding the test set and anonymizing usernames, and only SBERT increases by 0.5\% points in macro F1 (F1 = 32.7 vs 32.2). 

\begin{table}\renewcommand\arraystretch{1.2}
  \centering
  \begin{threeparttable}
  \scalebox{0.68}{
  \begin{tabular}{lccccccccccccccc}
    \hline
    & \multicolumn{3}{c}{Macro} && \multicolumn{3}{c}{Agree} && \multicolumn{3}{c}{Disagree} && \multicolumn{3}{c}{No Stance}\\
    \cline{2-4}\cline{6-8}\cline{10-12}\cline{14-16}
    & P & R & F1 && P & R & F1 && P & R & F1 && P & R & F1 \\
    \hline
    \multicolumn{16}{l}{\textit{Test on COVIDLies v0.1}} \\
    (orig) Linear, Bag-of-Words  & 35.2 & 38.1 & 24.0 && 9.8 & 59.7 & 16.9 && 10.5 & 28.9 & 15.4 && 85.3 & 25.8 & 39.7 \\
    (orig) Linear, Avg. GloVe & 35.9 & 40.8 & 26.6 && 15.8 & \textbf{68.5} & 25.7 && 4.2 & 21.6 & 7.1 && 87.5 & 32.2 & 47.1 \\
    (orig) SBERT & 36.1 & 40.1 & 32.2 && 17.6 & 31.9 & 22.7 && 6.1 & 37.6 & 10.5 && 84.7 & 50.6 & 63.4 \\ 
    (orig) SBERT (DA) & 51.1 & 47.3 & 41.5 && 58.1 & 23.4 & 33.4 && 8.7 & \textbf{50.4} & 14.9 && 86.5 & 67.9 & 76.1 \\
    (orig) BERTScore (DA) + SBERT (DA) & \textbf{55.9} & \textbf{50.9} & \textbf{50.2} && \textbf{63.3} & 30.6 & \textbf{41.2} && \textbf{14.4} & 34.1 & \textbf{20.3} && \textbf{90.0} & \textbf{88.0} & \textbf{89.0}\\
    \hline
    \multicolumn{16}{l}{\textit{Test on COVIDLies v0.2}} \\
    (repr) Linear, Bag-of-Words & 34.6 & 37.9 & 22.2 && 8.4 & 61.1 & 14.8 &&7.7 & 26.8 & 12.0 && 87.7 & 25.9 & 40.0 \\
    (repr) Linear, Avg. GloVe & 34.7 & 39.3 & 23.3 && 12.0 & \textbf{68.6} & 20.5 && 3.4 & 20.5 & 5.9 && 88.5 & 28.9 & 43.5\\
    (repr) SBERT & 36.6 & 41.3 & 32.7 && 17.1 & 33.5 & 22.6 && 5.0 & 36.6 & 8.8 && 87.7 & 53.7 & 66.6\\
    (repr) SBERT (DA) & 50.5 & 46.8 & 40.8 && 56.4 & 25.2 & 34.8 && 6.5 & \textbf{48.6} & 11.5 && 88.6 & 66.6 & 76.1\\
    (repr) BERTScore (DA) + SBERT (DA) & \textbf{54.8} & \textbf{51.8} & \textbf{50.1} && \textbf{59.9} & 32.2 & \textbf{41.9} && \textbf{12.4} & 34.7 & \textbf{18.3} && \textbf{92.1} & \textbf{88.4} & \textbf{90.2}\\
    \hline
  \end{tabular}}
  \end{threeparttable}
  \caption{Reproduction results of stance detection on COVIDLies. The prefix \textit{orig} indicates the results from the original study~\cite{hossain2020covidlies}, and the \textit{repr} indicates our reproduction results.} \label{lies_repr}
\end{table}

\section{Our Stance Detection Methods}~\label{methods} 
In this section, we describe the methods for our stance detection task, including model architectures and training strategies. We use two model architectures for stance detection: SBERT and Cross-Encoder. SBERT has shown to be effective for stance detection on the COVIDLies dataset. The model details are described in Section~\ref{sbert}. In the following part, we describe the architecture of Cross-Encoder and explain how it differs from SBERT. With the architecture of SBERT and Cross-Encoder, we use the models fine-tuned on the MNLI dataset, and then fine-tune them on the RumourEval or COVIDLies datasets. We explain this fine-tuning process in detail.

\subsection{Model Architectures}

\paragraph{BERT} Both SBERT and Cross-encoder use BERT as an encoder to obtain token representation. BERT~\cite{devlin-etal-2019-bert} is a transformer-based language model developed by Google for NLP tasks. It can represent each token based on its bidirectional context. This is due to the architecture of BERT, which is a multi-layer bidirectional Transformer encoders~\cite{vaswani2017attention}. The encoder uses a self-attention mechanism to learn word representations from both left and right contexts. BERT is pretrained on a large amount of BookCorpus and English Wikipedia using a combination of a masked language model objective (some tokens are randomly masked and the model predicts the words) and a next sentence prediction (given a pair of sentences, the model predicts if one sentence follows the other or not). Pretrained models carry abundant language knowledge and the model can be easily adapted to the target task by fine-tuning the model parameters using relatively small data from the target task. 

\paragraph{Cross-Encoder} For the BERT Cross-Encoder, a pair of sentences are passed to BERT simultaneously. The input starts with a special token [CLS], followed by the concatenation of a tweet and a misinformation item with the [SEP] token delimiting them. The [CLS] token embedding from the BERT output is used as the aggregate sequence representation for classification, since it is used to predict whether a sentence pair is coherent or not during pre-training. After BERT encoding, the [CLS] representation is fed into a fully-connected layer, which projects the hidden size into the label size. We use the Cross-Entropy loss as the optimization objective. Let $p(\hat{y}_i)$ be the probability distribution in three classes of the $i_{th}$ example, given by
\begin{equation}
    p(\hat{y}_i|h_i^c,\theta,\phi) = Softmax(\phi\cdot h_i^c)
\end{equation}
where $h_i^c$ is the [CLS] token representation from the BERT output of the $i_{th}$ example. $\theta$ are the BERT parameters, and $\phi$ are the parameters of the dense layer with a size of $m\times  k$, where $m$ is the dimension of BERT embeddings and $k$ is the number of labels. 
The Cross-Entropy loss $l$ is computed as follows:
\begin{equation}
        l(\theta,\phi) = -\sum_{i=1}^n \sum_{k=1}^K w_k\cdot p(\hat{y}_{i,k})\cdot y_{i,k}
\end{equation}
where $n$ is the number of examples in a mini-batch. $y_{i}$ is the label vector of the $i_{th}$ examples, which is 1 for the true class and 0 for the other classes. $w_k$ is the manual rescaling weight given to each class, which defaults to 1.

\begin{figure}
\centering
\begin{subfigure}[b]{.5\textwidth}
  \centering
  \includegraphics[width=.8\linewidth]{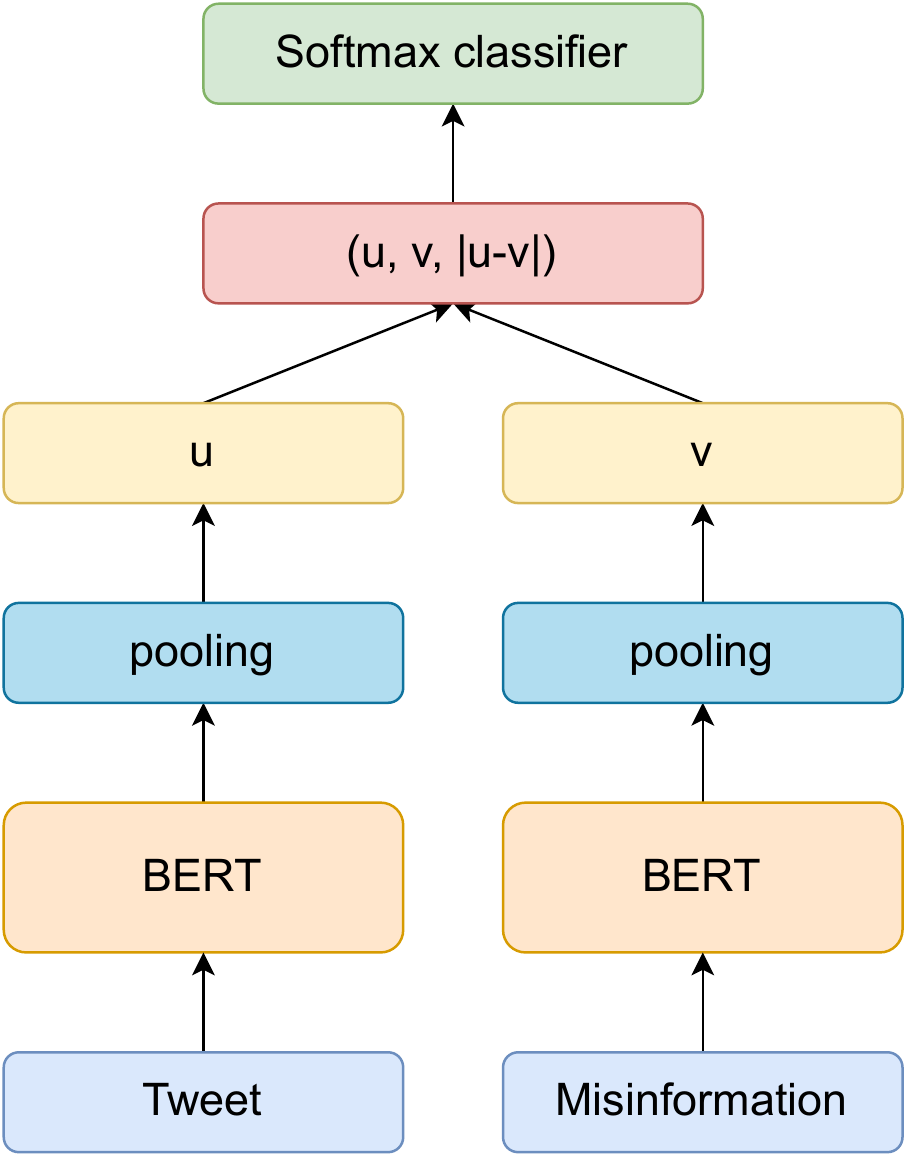}
  \caption{SBERT}
\end{subfigure}%
\begin{subfigure}[b]{.5\textwidth}
  \centering
  \includegraphics[width=.8\linewidth]{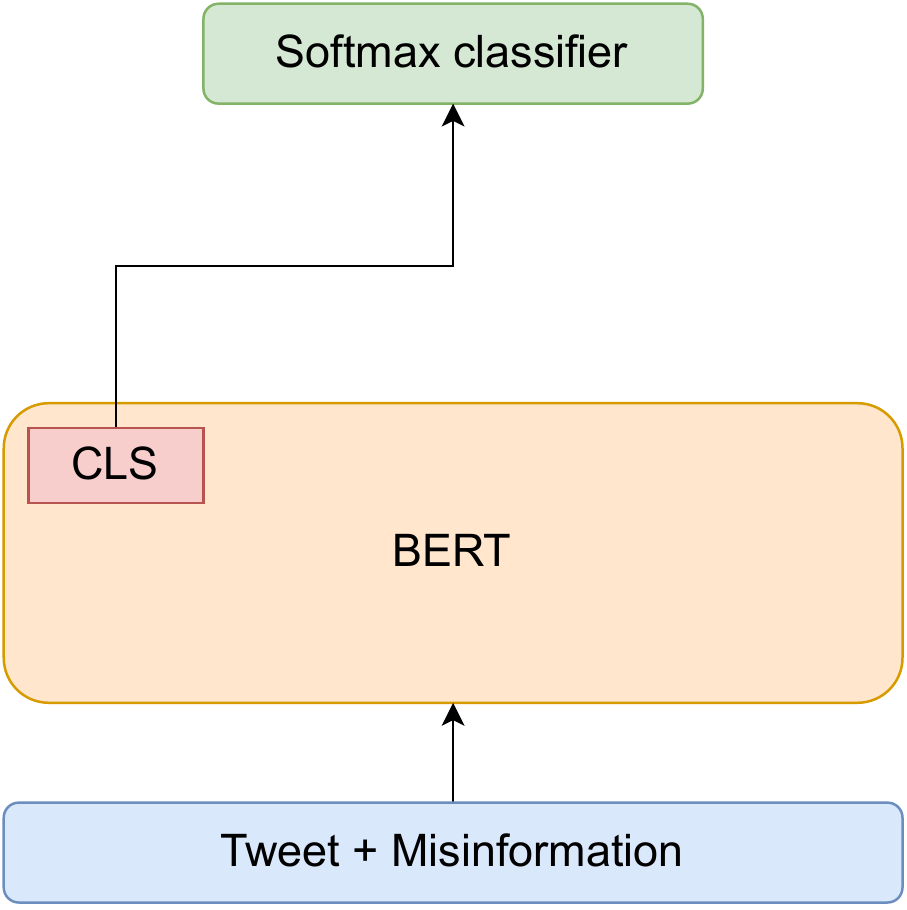}
  \caption{Cross-Encoder}
\end{subfigure}
\caption{SBERT (left), Cross-Encoder (right)}\label{models}
\end{figure}

\paragraph{Model comparison} 
Figure~\ref{models} shows the architectures of SBERT and Cross-Encoder. There are two main differences in the sentence-pair classification task between SBERT and Cross-Encoder: (1) Sentence representation: both sentences are fed into SBERT individually, while for Cross-Encoder, the concatenation of both sentences is passed to the network. Thus, SBERT can produce independent representation for each sentence, while Cross-Encoder cannot. (2) Token-level interaction: Cross-Encoder applies the attention to the tokens of both sentences across all transformer layers, while SBERT lacks the token-level interaction between two sentences.

\subsection{Training Strategies}\label{fine-tuning}
The model in our task is fine-tuned sequentially on the MNLI dataset and two stance detection datasets. The reason for fine-tuning in two stages is that these datasets are for different tasks and greatly differ in data size.

\paragraph{Similar task} NLI is a high-level understanding task that involves reasoning about the semantic relationships within sentences~\cite{conneau-etal-2017-supervised}. It is similar to our task and we could easily correspond our labels to NLI. Also, this task has much larger training datasets, which will help alleviate the problem of limited annotations. Therefore, we use the models fine-tuned on the MNLI dataset. We map \textit{Favor}, \textit{Against} and \textit{Neither} in our task to \textit{Entailment}, \textit{Contradiction} and \textit{Neutral} in NLI respectively.

\paragraph{Target domain}~\label{td} The NLI task is similar to ours but still in different domains, so we further fine-tune the models on two stance detection datasets, i.e. RumourEval and COVIDLies. Both datasets target rumors and have social media text. However, their class distributions are quite different from our dataset. Therefore, we adopt the following strategies for this problem:
\begin{itemize}
    \item Rescaling class weights~\cite{baziotis2017datastories}: This method takes the cost of prediction error into account. Specifically, the misclassification of the minority class will be penalized heavier than the majority class. The class weight $w_k$ is given by $w_k = max(x)/x_k$ , where $x$ is the vector with the class counts.
    \item Undersampling: we remove some examples from the majority classes of both datasets. Specifically, we keep all \textit{Against} examples and randomly select examples from \textit{Favor} and \textit{Neither} with a fixed probability, which is equal to the expected sample number divided by the class frequency. The expected sample number is 400 for COVIDLies and 550 for RumourEval. After undersampling, a combination of both balanced datasets is also used in our experiment. Table~\ref{undersample} shows the statistics of the balanced datasets after undersampling.
\end{itemize}

\begin{table}\renewcommand\arraystretch{1.2}
  \centering
  \begin{threeparttable}
  \scalebox{0.90}{
  \begin{tabular}{lccccc}
    \hline
    & Total & Favor & Against & Neither\\
    \hline
    RumourEval, balanced & 1607 & 554 (34.5\%) & 519 (32.3\%) & 534 (33.2\%) \\
    COVIDLies v0.2, balanced & 1185 & 416 (35.1\%) & 366 (30.9\%) & 403 (34.0\%) \\
    Combined dataset & 2792 & 970 (34.7\%) & 885 (31.7\%) & 937 (33.6\%) \\
    \hline
  \end{tabular}
  }
  \end{threeparttable}
  \caption{Statistics of undersampled RumourEval and COVIDLies. Includes the number and percentage of examples in each label.} \label{undersample}
\end{table}

\section{Experiments and Results}
In this section, we first experiment with how tweets and misinformation correspond to the premises and hypotheses in the NLI task. Second, we use the models fine-tuned on the MNLI dataset, and then fine-tune them on the RumourEval or COVIDLies datasets. We evaluate the model performance on our dataset. Finally, we do a dataset ablation study to identify their contributions to solving the stance detection task.

\paragraph{NLI models}\label{nli} Due to limited GPU resources, we did not fine-tune the models on the NLI dataset ourselves, but directly used the publicly available models that have been fine-tuned on the MNLI dataset. The two models are SBERT and Cross-Encoder, respectively.
\begin{itemize}
    \item \textit{mnli-sbert-ct}: This model has the architecture of SBERT and was released by Hossain et al.~\cite{hossain2020covidlies}. It is initialized with the COVID-Twitter-BERT weights and fine-tuned on the MNLI dataset. COVID-Twitter-BERT is a BERT-large model, further pretrained on a large corpus of COVID-19 related tweets ~\cite{muller2020covid}.
    \item \textit{mnli-bert-ct}: This model has the architecture of Cross-Encoder and was provided by Muller et al.~\cite{muller2020covid}. It uses the COVID-Twitter-BERT-v2 weights and is fine-tuned on the MNLI dataset. COVID-Twitter-BERT-v2 is identical to COVID-Twitter-BERT but trained on more data.
\end{itemize}

\paragraph{Tweet preprocessing} 
Before passing the sentences into the network, we process tweet texts according to the data preprocessing method used in COVID-Twitter-BERT~\cite{muller2020covid}.
\begin{itemize}
    \item Whitespace: Replace tab($\backslash$t), newline($\backslash$n) and carriage return ($\backslash$r) characters by spaces, and replace multiple spaces with a single space.
    \item Username: Replace username mentions with ``twitteruser". For multiple username mentions, prefix ``twitteruser" with the number. e.g. replace ``@user @user" with ``2 twitteruser".
    \item URL: Replace URL with ``twitterurl". For multiple URLs, prefix ``twitterurl" with the number. e.g. replace ``http://... http://.." with ``2 twitterurl".
    \item Emoji: Convert emojis into text aliases. e.g. the thumbs-up emoji becomes ``:thumbs\_up". This is implemented by an existing Python package called Emoji~\footnote{https://github.com/carpedm20/emoji/}.
\end{itemize}

\paragraph{Metrics} 
We use precision, recall, and F1 to measure the model performance, the same as in the COVIDLies paper~\cite{hossain2020covidlies}. The definitions of these metrics are described in Section~\ref{metric}. In addition, we add the accuracy metric, which is the percentage of correct predictions across all instances. 

\subsection{Sentence Mapping}

\paragraph{Experiment} We experimentally investigate how the tweet and misinformation in our task correspond to the premise and hypothesis in the NLI task. We apply two NLI models to the full RumourEval, COVIDLies and COVMis-Stance datasets with two mappings: (1) misinformation mapped to premises and tweets mapped to hypotheses; (2) misinformation mapped to hypotheses and tweets mapped to premises. The NLI models are \textit{mnli-sbert-ct} and \textit{mnli-bert-ct}.

\paragraph{Results} Table~\ref{sm} presents the results of different sentence mappings. The NLI models perform better on the stance detection datasets if we map tweets to premises and misinformation to hypotheses, rather than the reverse order. For RumourEval, this mapping performs about 2\% points higher in F1 score for both SBERT and Cross-Encoder. This improvement is more pronounced on the COVIDLies and COVMis-Stance datasets. For COVIDLies, the F1 score improves by 5\% points for SBERT and 7\% points for Cross-Encoder. For COVMis-Stance, the F1 score improves by 8\% points for SBERT and 9\% points for Cross-Encoder. 

We notice that the macro F1 of SBERT on COVIDLies is 2\% points lower than our reproduced result in Table~\ref{lies_repr}, under the condition that the same SBERT model and test set are used for both experiments (F1 = 38.83 vs 40.8). We found that this is because tweets are preprocessed in the sentence mapping experiment, while we did not do so in the reproduction experiment. The reason for the F1 drop after tweet preprocessing may be that the \textit{mnli-sbert-ct} model was originally proposed for stance detection in COVIDLies, and it may be validated on the COVIDLies dataset without tweet preprocessing.

\begin{table}\renewcommand\arraystretch{1.2}
  \centering
  \scalebox{0.85}{
  \begin{threeparttable}
  \begin{tabular}{lcccccccc}
    \hline
    &\multicolumn{2}{c}{RumourEval} && \multicolumn{2}{c}{COVIDLies} && \multicolumn{2}{c}{COVMis-Stance} \\
    \cline{2-3}\cline{5-6}\cline{8-9}
    & SBERT & Cross-Encoder && SBERT & Cross-Encoder && SBERT & Cross-Encoder \\
    \hline
    p-mis, h-tweet & 39.55 & 41.40 && 33.53 & 31.64 && 30.21 & 22.57 \\
    p-tweet, h-mis & 41.73 & 43.37 && 38.83 & 38.86 && 38.67 & 41.27 \\
    \hline
  \end{tabular}
  \caption{Results of two sentence mappings from NLI to our task. We present the macro F1 of the NLI models on the full RumourEval, COVIDLies, and COVMis-Stance datasets. \textit{p}, \textit{h} and \textit{mis} refer to premise, hypothesis and misinformation respectively.} \label{sm}
\end{threeparttable}}
\end{table}

\subsection{Stance Detection}

\paragraph{Experiments}\label{exp} 
We fine-tune the NLI models on each of the following five datasets and evaluate the model performance on our dataset, which is randomly divided into a validation set and a test set in a ratio of 2:8. We use the validation set for parameter tuning and the test set for evaluation. 
\begin{itemize}
    \item RumourEval (full): To alleviate the class imbalance problem in this dataset, we pass to the loss function the class weights described in Section~\ref{td}.
    \item RumourEval (balanced): The model is fine-tuned on the balanced RumourEval, which is the result of undersampling on the majority classes as described in Section~\ref{td}.
    \item COVIDLies (full): The class weights are also applied when fine-tuning the model on the full COVIDLies dataset, so that the model can adjust to the class imbalance.
    \item COVDILies (balanced): The balanced COVIDLies dataset as described in Section~\ref{td}.
    \item Combined dataset: A combination of the balanced RumourEval and COVIDLies datasets to increase the diversity and number of training instances. 
\end{itemize}
 
\paragraph{Configurations} The parameter settings are described as follows: 
\begin{itemize}
    \item Maximum input sequence length: The input sequence is padded or truncated to 128 tokens for SBERT and 256 tokens for Cross-Encoder. This refers to the SBERT setting in the COVIDLies paper~\cite{hossain2020covidlies}. We also analyze sentence lengths of our experimental datasets. All sentences are within 61 words for COVIDLies and within 60 words for the COVMis-Stance dataset. For RumourEval, about 1\% of sentences exceed 128 words. Thus, limiting the sequence length to 128 or 256 tokens should have little impact on model performance, and also be computationally efficient.
    \item Loss function: Cross-Entropy loss. The class weights are 1 by default. They are rescaled for the full RumourEval or COVIDLies dataset.
    \item Optimizer: We use Adam optimizer with the learning rate of $2\times 10^{-5}$, L2 weight decay of 0.01, learning rate warm-up over the first 10\% of training steps, and linear decay of the learning rate. Most of these parameters follow the BERT fine-tuning settings for GLUE tasks~\cite{devlin-etal-2019-bert}, except for the warm-up steps, which equals 10,000 in the BERT paper. As we have fewer training steps, we set this value as a ratio rather than an exact number of steps.
    \item Batch size: Number of samples processed before the model updating. The batch size is set to be 8, which depends on the memory size of our GPU.
    \item Number of epochs: Number of complete passes through the training dataset. The epoch is set to be 3, referring to the BERT paper~\cite{devlin-etal-2019-bert}. 
\end{itemize}

\paragraph {Stance detection results} Table~\ref{ours_test} presents the stance detection results of all models on our test set. The first lines of SBERT and Cross-Encoder show the results of the NLI models, which are used as the baseline results in our experiments. We make the following observations.

\begin{itemize}
    \item \textit{Best model}: The Cross-Encoder fine-tuned on the combined dataset achieves the best result in terms of macro F1. It outperforms the Cross-Encoder baseline by 16\% points in macro F1 (F1 = 57.28 vs 41.83).

    \item \textit{Fine-tuning}: Fine-tuning the NLI models on RumourEval or COVIDLies improves the model performance but with two exceptions. Both of them are SBERT and fine-tuned on the full but unbalanced datasets. These two datasets are the full RumourEval and COVIDLies datasets. They have a slight decrease in macro F1 compared to the SBERT baseline (F1 = 39.30 for RumourEval, F1 = 39.38 for COVIDLies).
    
    \item \textit{SBERT vs Cross-Encoder}: Overall, the Cross-Encoders outperform the SBERT models with the same data fine-tuning. They perform about 2\%-5\% points higher in macro F1 than SBERT. This is because the two sentences interact across all transformer layers of Cross-Encoder, while for SBERT, they do not have any interactions before the classifier. From a category perspective, the Cross-Encoders perform better on the \textit{Favor} and \textit{Against} classes than SBERT, but do not improve on the \textit{Neither} class. Specifically, the Cross-Encoders performs an average 5\% points F1 higher than SBERT on the \textit{Favor} class, if excluding the full RumourEval model, which shows a 30\% points difference in F1 (F1 = 70.12 vs 40.19). Also, it performs an average 10\% points F1 higher than SBERT on the \textit{Against} class. However, the Cross-Encoders have a considerable drop in recall for the \textit{Neither} class, except for the full COVIDLies model (R = 70.99 vs 70.23). This indicates that the Cross-Encoders predict fewer examples to be \textit{Neither} than SBERT.
    
    \item \textit{Class imbalance}: We experimented with two strategies to solve the class imbalance problems of RumourEval and COVIDLies: (1) fine-tuning on the full datasets but rescaling class weights; (2) fine-tuning on the balanced datasets obtained by undersampling. The results show that the models with the second strategy outperform those with the first strategy. The SBERT model with rescaled class weights performs even worse than the baseline (F1 = 39.30 for RumourEval, F1 = 39.38 for COVIDLies), while the model using the balanced dataset shows a sizable improvement (F1 = 54.39 for RumourEval, F1 = 46.32 for COVIDLies). The Cross-Encoders using either of the two strategies perform better than the baseline, but they show greater improvement with the undersampling strategy (F1 = 54.10 vs 57.06 for RumourEval, F1 = 49.62 vs 52.51 for COVIDLies).

    \item \textit{Class performance}: All models perform the worst on the \textit{Neither} class, with the highest precision only reaching 21.56\% points. This indicates that the models incorrectly predict many examples as \textit{Neither}. On the other hand, most models obtain higher F1 scores on the \textit{Favor} class than \textit{Against}, except for the SBERT with the full RumourEval dataset (\textit{Favor} F1 = 40.19, \textit{Against} F1 = 50.57) and the Cross-Encoder with the balanced RumourEval dataset (F1 = 70.97 for \textit{Favor}, F1 = 73.61 for \textit{Against}). 
    
    \item \textit{RumourEval vs COVIDLies}: We compare the performance of the models fine-tuned on the balanced RumourEval or COVIDLies datasets. RumourEval outperforms COVIDLies by 8\% points F1 for SBERT  (F1 = 54.39 vs 46.32) and 5\% points F1 for Cross-Encoder (F1 = 57.06 vs 52.51), because of its better performance on the \textit{Against} class. Specifically, RumourEval performs a 21\% points F1 higher than COVIDLies on the \textit{Against} class for SBERT (F1 = 62.20 vs 41.37) and a 17\% points higher for Cross-Encoder (F1 = 73.61 vs 56.02). For the \textit{Favor} class, we observe that the Cross-Encoder using COVIDLies performs a 6\% points higher in F1 than the one using RumourEval (F1 = 76.43 vs 70.97 ). However, they show a minor difference in the \textit{Neither} class.
    
\end{itemize}

\begin{table}\renewcommand\arraystretch{1.2}
  \centering
  \begin{threeparttable}
  \scalebox{0.68}{
  \begin{tabular}{lccccccccccccccc}
    \hline
    & \multicolumn{3}{c}{Favor} && \multicolumn{3}{c}{Against} && \multicolumn{3}{c}{Neither} && \multicolumn{3}{c}{Macro}\\
    \cline{2-4}\cline{6-8}\cline{10-12}\cline{14-16}
    & P & R & F1 && P & R & F1 && P & R & F1 && P & R & F1\\
    \hline
    \multicolumn{16}{l}{\textit{Sentence-BERT}}\\
    mnli & 82.75  & 37.75  & 51.85  && 64.20  & 31.03  & 41.83  && 15.31  & 72.52  & 25.28  && 54.09  & 47.10  & 39.65 \\
    +rumoureval, full & \textbf{93.45}  & 25.60  & 40.19  && 78.79  & 37.23  & 50.57  && 16.05  & \textbf{87.79}  & 27.14  && 62.76  & 50.21  & 39.30 \\
    +rumoureval, balanced & 76.35  & 64.94  & 70.18  && 74.83  & 53.22  & 62.20  && \textbf{21.56}  & 53.82  & \textbf{30.79}  && 57.58  & 57.33  & 54.39 \\
    +covidlies, full & 72.91  & 61.65  & 66.81  && 71.18  & 14.44  & 24.01  && 16.96  & 70.23  & 27.32  && 53.68  & 48.77  & 39.38 \\
    +covidlies, balanced & 78.47  & 62.45  & 69.55  && 68.70  & 29.59  & 41.37  && 17.90  & 64.50  & 28.03  && 55.02  & 52.18  & 46.32 \\
    +combined & 82.86 & 61.65 & 70.70 && 72.54 & 51.07 & 59.94 && 20.21 & 59.16 & 30.13 && 58.54 & 57.29 & 53.59 \\
    \multicolumn{16}{l}{\textit{Cross-Encoder}}\\
    mnli & 80.34  & 41.93  & 55.10  && 78.22  & 32.58  & 46.00  && 14.78  & 69.47  & 24.38  && 57.78  & 47.99  & 41.83 \\
    +rumoureval, full & 84.22  & 60.06  & 70.12  && 88.20  & 47.26  & 61.54  && 19.60  & 70.23  & 30.64  && \textbf{64.01}  & \textbf{59.18}  & 54.10 \\
    +rumoureval, balanced$^*$ & 80.81  & 63.65  & 70.97  && 81.50  & \textbf{67.33}  & \textbf{73.61}  && 19.00  & 44.66  & 26.61  && 60.43  & 58.54  & 57.06 \\
    +covidlies, full & 80.57  & 67.73  & 73.59  && \textbf{88.50}  & 30.31  & 45.16  && 19.12  & 70.99  & 30.12  && 62.73  & 56.34  & 49.62 \\
    +covidlies, balanced$^*$ & 76.36  & \textbf{76.51}  & 76.43  && 81.59  & 42.70  & 56.02  && 17.54  & 44.04  & 25.08  && 58.50  & 54.42  & 52.51 \\
    +combined$^*$ & 78.62  & 75.66  & \textbf{77.06}  && 80.14  & 60.43  & 68.62  && 20.10  & 38.17  & 26.16  && 59.62  & 58.09  & \textbf{57.28} \\
    \hline
  \end{tabular}}
  \end{threeparttable}
  \caption{Results of stance detection on our test set. Precision (P), Recall (R) and F1 are presented for each class as well as the macro averages values. The model name with * means that we run the model five times with different random seeds and each cell value in this row is the average score over five times.} \label{ours_test}
\end{table}

\paragraph{Comparison between queries} Table~\ref{source_res} shows the results of the Cross-Encoders for different query types as well as a majority class baseline (each instance is classified into the majority class). The blank cells in the table indicates that the macro F1 scores of the corresponding models are not available, because there is one class without predicted instances. Overall, the examples retrieved by keywords are relatively easier to predict than those retrieved by URLs. As the macro F1 for the title examples cannot be obtained, we do not compare the model performance between titles and the other two queries.

All models perform a high accuracy on the examples retrieved by titles, but it cannot reflect the predictive ability of the models, since nearly 83\% of title examples are labeled as \textit{Favor}. However, the models still perform better than the majority class baseline and the COVIDLies model is the most accurate.

For the examples sourced by URL retrieval, all models have a lower accuracy than the majority class baseline. Among them, the RumourEval model has the highest accuracy but a 0.7\% points lower in F1 than the model with the combined dataset. The COVIDLies model performs much worse than the other two models, with an 18\% points lower in accuracy than the RumourEval model and a 10\% points lower in F1.

For the examples sourced by keywords retrieval, all models outperform the majority class baseline. The COVIDLies model performs the best, closely followed by the model with the combined dataset. The RumourEval model performs the worst, with a 7\% points lower in accuracy than the COVIDLies model and 1\% point lower in F1.

From the above results, we see that the RumourEval models perform better on the URL and \textit{Against} examples, while the COVIDLies models perform better on the keywords and \textit{Favor} examples. Their performance for queries is consistent with them for classes, since 59\% of URL examples are marked as \textit{Against} and 57\% of keywords examples are marked as \textit{Favor}. This indicates that the URL examples are more similar to RumourEval, probably because both are responses to news articles or source tweets. Some of them are simply comments on the rumor and do not repeat its content. In contrast, the keyword examples usually mention the content of the rumor. They are similar to the COVIDLies examples, because the tweets in COVIDLies are selected based on their similarity to the rumor.

\begin{table}\renewcommand\arraystretch{1.2}
  \centering
  \begin{threeparttable}
  \scalebox{0.9}{
  \begin{tabular}{lcccccccc}
    \hline
     & \multicolumn{2}{c}{Title} && \multicolumn{2}{c}{URL} && \multicolumn{2}{c}{Keywords} \\
    \cline{2-3}\cline{5-6}\cline{8-9}
    & Acc & F1 && Acc & F1 && Acc & F1\\
    \hline
    Majority class baseline & 82.62 & - && 59.48 & - && 56.91 & - \\
    \hline
    % \multicolumn{9}{l}{\textit{Based on the NLI model}} \\
    Rumoureval, balanced $^*$ & 84.09 & 56.86 && \textbf{55.58} & 48.14 && 64.86 & 57.95 \\
    COVIDLies, balanced $^*$ & \textbf{89.46} & - && 37.55 & 38.25 && \textbf{71.21} & \textbf{59.01}\\
    Combined $^*$ & 88.92 & - && 52.43 & \textbf{48.86} && 70.93 & 58.63 \\
    \hline
  \end{tabular}}
  \end{threeparttable}
  \caption{Results for different queries in terms of accuracy and macro F1. Includes the results of the majority class baseline and three Cross-Encoders fine-tuned on different datasets. The model name with * means that we run the model five times with different random seeds and each cell value in this row is the mean score of five times.}  \label{source_res}
\end{table}

\subsection{Ablation of Datasets}
\paragraph{Experiments} 
We conduct an ablation study across three datasets: MNLI, RumourEval (balanced), and COVIDLies (balanced). To quantify their contributions, we perform two experiments: (1) we only keep one dataset for fine-tuning; (2) we remove one dataset and use the remaining two datasets for fine-tuning. All models in this study are Cross-Encoders.

\paragraph{Results}
Table~\ref{ablation} shows the results of our dataset ablation study. The results of the only-one experiment are not fully consistent with those of the all-without-one experiment. In the only-one experiment, COVIDLies performs the best, slightly better than MNLI, and RumourEval ranks last. However, in the all-without-one experiment, COVIDLies has the least impact, while MNLI has the biggest impact, followed by RumourEval. 

The reason for the good performance of only COVIDLies may be that it predicts well on the examples retrieved by keywords, which account for the majority of our dataset. However, there is an overlap between COVIDLies and the other two datasets in terms of example types, since both MNLI and RumourEval have examples in which two sentences are semantically similar. This might result in the least drop in model performance after excluding COVIDLies. The exclusion of RumourEval has a greater impact on model performance than that of COVIDLies, probably because both MNLI and COVIDLies lack instances similar to URL examples. Thus the model ability in this aspect is weakened without RumourEval.

MNLI performs closely to COVIDLies in the only-one experiment, but it has about 330 times the number of training instances as COVIDLies. This suggests that a sufficient number of NLI instances compensate for the discrepancies in tasks and domains to a certain extent. After excluding NLI, the training set becomes much smaller and the model has a 9\% points drop in macro F1. 

Only RumourEval or COVIDLies show high standard deviations. Combining the data sizes of RumourEval (1607) and COVIDLies (1185), we speculate that a relatively small dataset may be insufficient for large model training. However, the standard deviations decrease considerably when the model is fine-tuned on MNLI before, indicating the MNLI fine-tuning enhances the model's stability.

\begin{table}\renewcommand\arraystretch{1.2}
  \centering
  \begin{threeparttable}
  \scalebox{0.85}{
  \begin{tabular}{lccccccccc}
    \hline
    & \multicolumn{3}{c}{Only} && \multicolumn{3}{c}{All without} && All \\
    \cline{2-4}\cline{6-8}\cline{10-10}
    & MNLI & RumourEval & COVIDLies && -MNLI & -RumourEval & -COVIDLies && \\
    \hline
    Acc & 41.63  & 41.40  & 46.84  && 55.99  & 59.00  & 62.75  && 64.93 \\
    SD & / & 3.30  & 6.09  && 1.36  & 1.56  & 1.77  && 2.18 \\
    Macro F1 & 41.83  & 37.18  & 42.06  && 48.23  & 52.51  & 57.06  && 57.28 \\
    SD & / & 3.18  & 4.01  && 1.81  & 1.21  & 1.23  && 1.63 \\
    \hline
  \end{tabular}
  }
  \end{threeparttable}
  \caption{Results of dataset ablation study. We run each model five times with different random seeds and present the average values of accuracy and macro F1 as well as the standard deviation (\textit{SD}).}\label{ablation}
\end{table}

\subsection{Error Analysis}
Table~\ref{error} presents some typical examples misclassified by the Cross-Encoder fine-tuned on the combined dataset, We analyze each example, as follows.

\begin{itemize}
    \item The model incorrectly predicts \textit{Neither} as \textit{Favor}. The tweet has a high degree of lexical overlap with the misinformation, but it is irrelevant. 
    \item The model incorrectly predicts \textit{Against} as \textit{Favor}. The tweet rewrites the misinformation, but it disagrees with the misinformation. 
    \item The tweet shares the fake news article and expresses a supportive view. It is not literally similar to the rumor, which might lead the model to incorrectly predicts it as \textit{Neither}.
    \item The last example is a reply tweet. It seems to refute the rumor, but not quite sure. We found this tweet is retrieved by the URL of a fact-checking article. Also, by checking the contextual conversation, we confirm that it should be flagged as \textit{Against}. Such cases are difficult for the models due to insufficient information, although the tweet text has some indications.
\end{itemize}

\begin{table}\renewcommand\arraystretch{1.2}
  \centering
  \begin{threeparttable}
  \scalebox{0.8}{
  \begin{tabular}{p{1.5cm}p{1.5cm}p{15cm}}
    \hline
    Label & Pred & Example \\
    \hline
    Neither & Favor & \textbf{Misinformation:} Flu and pneumonia vaccines can protect against COVID-19.\newline \textbf{Tweet:} HEALTH IS WEALTH. We got covered! Today, me and my family got our flu and pneumonia vaccine shots for protection against flu and pneumonia infections.  We hope to get the covid 19 vaccines, too when available.… https://t.co/h3FLCGDcND\\
    \hline
    Against & Favor & \textbf{Misinformation:} CDC does not recommend that people who are healthy wear face masks to protect themselves from respiratory diseases, including COVID-19. \newline \textbf{Tweet:} @username @username @username The CDC is incompetent as well. Last week they were telling us healthy people don't need mask even though people can be asymptomatic carriers. I recommend you look up Hong Kong and South Korea's CDC guidelines instead \\
    \hline
    Favor & Neither & \textbf{Misinformation:} The coronavirus is a military bioweapon developed by China's army. \newline \textbf{Tweet:} Everybody need to read this and decide whether or not it’s the truth cause I believe it! https://t.co/iJNgMNNMUS\\
    \hline
    Against & Neither & \textbf{Misinformation:} Being exposed to the sun for two hours kills the 2019 coronavirus. \newline \textbf{Tweet:} @username @username Not true. But sunshine might make you feel better. https://t.co/vSZDgXdXhJ \\
    \hline
  \end{tabular}}
  \end{threeparttable}
  \caption{Examples of misclassification}\label{error}
\end{table}

\section{Discussion}

In this section, we discuss what improvements can be made from five perspectives: data construction, solutions to the examples retrieved by different queries, class imbalance, multi-dataset learning, and differences between SBERT and Cross-Encoder.

\paragraph{Data construction}
We constructed the COVMis-Stance dataset through data collection, sampling, and annotation. There are several aspects that could be improved in this process.

\begin{itemize}
    \item Selection of misinformation items: We select the misinformation items based on the number of related tweets retrieved by news titles and URLs, as this number reflects the interest level on this topic to some extent. However, it sometimes deviates from the actual interest level, because it does not take into account the related tweets that do not share URLs but discuss the same topics. Also, the tweeters might share news articles because of other events rather than the corresponding rumor. On top of that, we observe that the political-related rumors rank relatively high with this method. This might limit the diversity of COVID-19 misinformation items. For such rumors, people's bias towards political figures or parties influences the stance on rumors. This also cannot be used as supporting information for determining the rumor veracity. Therefore, only considering the number of relevant tweets might be insufficient for misinformation selection, and improving the quality of misinformation items might help the study.
    
    \item Description of misinformation items: Some misinformation items are not the same as the titles of news articles or fact-checking articles, since we do simple processing for them. However, we expect such processing brings minor improvement since pretrained language models are usually robust to such text noise. In addition, this manual checking takes time as the number of misinformation items increases. Therefore, we can use news headlines as our target without any manual intervention. This also can be used to verify the model's robustness.
    
    \item Relevant tweets: We link tweets to COVID-19 rumors through URLs and keywords, but manual construction of keywords will be infeasible with an increasing number of misinformation items. An extended idea is to automate the construction of keywords. Besides, Hossain et al.~\cite{hossain2020covidlies} address it by regarding it as a misinformation retrieval task. Given a tweet, they select the most relevant misinformation item from the pre-defined misinformation items based on sentence similarity. Another idea is to first cluster the tweets that are discussing the same news into a group and then associate them with specific misinformation items. 
    
    \item \textit{Neither} class: We group the examples that neither support nor refute the rumors into one category, thus the \textit{Neither} class is complicated. Specifically, the examples in the \textit{Neither} class might have different stance strengths. For example, some tweets question the rumor veracity, while some tweets are neutral to them. In addition, they may have different relevance. The examples above are discussing the corresponding rumors, while some examples may have nothing to do with the rumors. Therefore, it may be helpful to divide the \textit{Neither} class into two or three groups. Meanwhile, it needs a number of training instances for these classes.
\end{itemize}

\paragraph{Query differences} 
The model performs differently for the examples retrieved by different queries. We illustrate the characteristics of these examples and discuss some possible improvements.

\begin{itemize}
    \item URL: The tweets share news articles and have comments from users. We found some of them hard to understand, because they mention the event details or respond to other events covered in news articles. For such cases, the news body text might be useful, since it helps understand the event to which the tweets are referring and provide additional information about the event.
    \item Keywords: Compared to URLs, the tweet retrieved by keywords is usually a complete statement. For such instances, we can frame it as a semantic matching problem, i.e., whether two sentences express the same meaning. The NLI and the datasets on semantic similarity are good choices to fine-tune the models.
    \item Reply: We see replying tweets via Twitter search, so we do not exclude this part from the dataset. However, putting such replies in conversational contexts may give a better understanding,
\end{itemize}

\paragraph{Class imbalance} The severe class imbalance exists in the RumourEval and COVIDLies datasets. We reconstruct a balanced dataset by undersampling the majority classes. The results show that this method is more effective than rescaling class weights in our task. However, undersampling reduces the number of training instances drastically, which has a negative effect on the training of large models. Oversampling the minority class is another solution to this problem, as it will guarantee all available training data is leveraged.

\paragraph{Multi-dataset learning} We fine-tune the model on the MNLI dataset and two stance detection datasets sequentially. We discuss possible improvements for the second phase, where the RumourEval and COVIDLies datasets are directly mixed to fine-tune the whole model.
\begin{itemize}
    \item RumourEval and COVIDLies perform quite differently in the examples obtained by different query types. Therefore, instead of directly mixing them, other combinations of datasets might be worth experimenting with, such as the method used in~\cite{schiller2021stance}, in which multiple datasets share the encoder but have their dataset-specific layers.
    
    \item The model learns useful capabilities from the MNLI fine-tuning, as can be seen from the result that the model with only MNLI performs closely to that with only COVIDLies. Therefore, in the context where the undersampled RumourEval or COVIDLies datasets are not large, we have the question of whether fine-tuning the whole model is an optimal choice. Parameter-efficient fine-tuning methods might be another option, such as selecting a small number of weights to update or fine-tuning a separate, small network that is tightly coupled with the model~\cite{min2021recent}. 
\end{itemize}

\paragraph{SBERT vs Cross-Encoder} 
SBERT underperforms Cross-Encoder in our task. The result is expected since two sentences have fewer interactions in SBERT than in Cross-Encoder. However, SBERT is more computationally efficient than Cross-Encoder, since it produces independent sentence representations. When the data becomes large, we could obtain the sentence representation of each misinformation item in advance and store them. When a tweet comes, we only need to calculate the sentence representation of the tweet by deep networks and combine it with the pre-computed sentence representation of the misinformation item for classification. In contrast, Cross-Encoder has to compute the vector representation for each combination of misinformation item and tweet, which will be much slower. Recently, many studies train a Cross-Encoder and distill the knowledge it learns into SBERT for computation efficiency. 

\section{Conclusions}

In this work, we studied stance detection on Twitter towards COVID-19 misinformation. We constructed a stance dataset, consisting of 2631 tweets with their stance towards COVID-19 misinformation. In addition to the dataset, we establish stance detection models with the SBERT and Cross-Encoder architectures. The results show that Cross-Encoder outperforms SBERT in terms of precision, recall, and F1. 

We leverage the BERT models fine-tuned on the NLI dataset. The model performance considerably drops without it, indicating that using the data-rich NLI as an intermediate task improves the performance of stance detection. Furthermore, the sentence correspondence from NLI to stance detection has a great impact on the model performance. It achieves better results when the tweet and misinformation are mapped to the premise and hypothesis respectively.

We fine-tune the NLI models on the RumourEval and COVIDLies datasets. Both datasets are heavily unbalanced in class distribution, and such difference from our data negatively influences the model performance. The experimental results show that undersampling is more effective than rescaling class weights in resolving this problem in our task. We found that the RumourEval model is better at predicting the examples sourced by URL retrieval, and the COVIDLies model performs better on the examples sourced by keywords retrieval. The Cross-Encoder fine-tuned on the mixture of RumourEval and COVIDLies combines the advantages of both datasets, achieving the best results among all models. 

For future work, we recommend adopting different methods for the examples retrieved by different query types. Other multi-dataset learning methods should also be explored. In addition, we published our dataset to facilitate cross-dataset evaluation of related studies, i.e., training and testing using different datasets.

%Bibliography
%\bibliographystyle{unsrt}  
\bibliographystyle{plain}  
%\bibliography{references}  

\clearpage
\appendix
\section{Appendix}~\label{appendix}
\subsection{Construction of Misinformation Items}~\label{mc}

Each misinformation item is described as a sentence based on titles of fake news or fact-checking articles. If both titles are provided by CoAID, we manually choose the simpler and clearer one. For example, a fact-checking article has the title of \textit{``The RT-PCR test for the virus that causes COVID-19 detects human DNA on chromosome 8, therefore all tests will give a positive result"}, and the corresponding news article has the title of \textit{``WHO Coronavirus PCR Test Primer Sequence is Found in All Human DNA"}. We use the title of the news article, because it is simpler and does not involve the raw word ``chromosome". In addition, we process the descriptions of misinformation items as follows.

\begin{itemize}
    \item \textbf{Split} Some rumors contain several inaccurate points and discussions about these points are also distributed, so we manually split it into multiple items. 

    \item \textbf{Merge} Some misinformation items are from the same news but might be reported by different media or checked by different websites. For such cases, we merge them into one message. 
    
    \item \textbf{Correct} Some titles from fact-checking websites are not rumors themselves but corrected information, like the title \textit{``No evidence that 5G is being forcibly installed in schools"} is the clarification of fake news. Therefore, we change it into \textit{``5G is being forcibly installed in schools"}.
    
    \item \textbf{Simplify} Some titles are the exact words spoken by some people and we summarize them into one sentence to make them clear and complete. For example, for the title \textit{``The U.S. went from 75,000 flu deaths last year in America to almost 0; are there allocation games being played to manipulate the truth ?"}, we rewrite it into \textit{``Trump claims that flu deaths in America are down to almost zero and data is being manipulated"}.
\end{itemize}

\subsection{Keywords Extraction}\label{ke}
The keywords extraction follows the following principles. In addition, we also test the keywords on Twitter's search interface to obtain as many relevant tweets as possible.
\begin{itemize}
    \item We first use relevant entity names as queries. For example, for the rumor \textit{``Shanghai government officially recommends Vitamin C for COVID-19"}, the keywords are \textit{Shanghai}, \textit{Vitamin C}, and \textit{COVID-19}. 
    \item If the claims are said by famous people, we usually add the individual's name to make it specific. For example, the keywords of \textit{``Nobel laureate Luc Montagnier claimed that the coronavirus genome contained sequences of HIV (the virus that causes AIDS)."} are \textit{Luc Montagnier}, \textit{coronavirus} and \textit{HIV}".
    \item Some verbs convey the important information in a sentence, indicating the relationship between two entities or the object's action. For the sentence \textit{``Trump administration refused to get coronavirus testing kits from the WHO"}, the keywords are \textit{Trump}, \textit{refused} and \textit{WHO test kits}. 
    \item Sometimes the keywords based on the above strategies still give a large number of irrelevant tweets. For such cases, we directly use a sentence as the query. For example, the sentence \textit{``Quotes Joe Biden as saying people who have never died before are now dying from coronavirus."} are retrieved by the query \textit{``people who have never died before are now dying from coronavirus"}.
\end{itemize}

\subsection{Implementation of Data Sampling}\label{ids}
According to the sampling strategy described in Section~\ref{sampling}, each example will be assigned a probability of being selected. Specifically, for each query type of each misinformation item, if the number of relevant tweets does not exceed 6, then the chosen probability is 1; otherwise, the probability is the sum of two items. One is the probability of being selected the first time $p_1=6/N$, where $N$ is the number of tweets of this query type; the other is the probability of being selected the second time $p_2=(1-p_1)\cdot m/N_r$, where $m$ is the number of tweets stilled needed to reach 24, and $N_r$ is the number of remaining tweets for this misinformation item. After determining the selected probability for each example, if a random number between 0.0 and 1.0 is less than the probability, then this example will be selected for annotation; otherwise this will not be selected.
\end{document}